\documentclass{article}

\usepackage[bottom]{footmisc}
\usepackage{float}
\usepackage{sidecap}
\usepackage{multirow}
\usepackage{parskip}

\usepackage[margin=1.25in]{geometry}
\usepackage[utf8]{inputenc} % allow utf-8 input
\usepackage[T1]{fontenc}    % use 8-bit T1 fonts
\usepackage{url}            % simple URL typesetting
\usepackage{booktabs}       % professional-quality tables
\usepackage{amsfonts}       % blackboard math symbols
\usepackage{nicefrac}       % compact symbols for 1/2, etc.
\usepackage{microtype}      % microtypography
\usepackage{kpfonts}
\usepackage{graphicx}
\usepackage{multicol}
\usepackage{enumitem}
\usepackage{wrapfig}
\usepackage{natbib}
\usepackage{smile}

\usepackage[small]{caption}
\usepackage[table]{xcolor}
\definecolor{codeblue}{rgb}{0,0.28,0.67}

\usepackage{hyperref}
\hypersetup{
    colorlinks,
    linkcolor={codeblue},
    citecolor={codeblue},
    urlcolor={codeblue}
}

\usepackage[scaled]{beramono}

\usepackage{lipsum}
\usepackage{listings}

\lstdefinestyle{mystyle}{
    escapechar=\%,
    keywordstyle=\color{codeblue},
    basicstyle=\ttfamily\small,
    breakatwhitespace=false,         
    breaklines=true,                 
    captionpos=b,                    
    keepspaces=true,                 
    numbers=left,                    
    numbersep=5pt,                  
    showspaces=false,                
    showstringspaces=false,
    showtabs=false,                  
    tabsize=2
}
\usepackage{amsthm}
\usepackage{mathtools}

\usepackage{setspace}

\def\ie{\textit{i.e.}}
\def\eg{\textit{e.g.}}

\newcommand{\myparagraph}[1]{\vspace{1pt}\noindent{\bf{#1}}~~}
\makeatletter
\renewcommand{\paragraph}{%
  \@startsection{paragraph}{4}%
  {\z@}{0em}{-1em}%
  {\normalfont\normalsize\bfseries}%
}
\makeatother

\usepackage{xspace}

\newcommand{\ce}{\textnormal{CE}\xspace}

\newcommand{\acc}{\textnormal{acc}}

\title{\Large
{Uncovering Memorization Effect in the Presence of Spurious Correlations}
\vspace{1.5ex}
}
\author{
\normalsize
Chenyu You$^{*,\dagger,1,6,7}$,
Haocheng Dai$^{*2}$, 
Yifei Min$^{*3}$, \\
\normalsize
Jasjeet S. Sekhon$^3$,
Sarang Joshi$^2$, 
James S. Duncan$^{1,4,5}$\\
\\
\normalsize $^{1}$ Department of Electrical \& Computer Engineering, Yale University, New Haven, CT, USA\\
\normalsize $^{2}$ Scientific Computing and Imaging Institute, University of Utah, Salt Lake City, UT, USA\\
\normalsize $^{3}$ Department of Statistics and Data Science, Yale University, New Haven, CT, USA\\
\normalsize $^{4}$ Department of Biomedical Engineering, Yale University, New Haven, CT, USA\\
\normalsize $^{5}$ Department of Radiology \& Biomedical Imaging, Yale University, New Haven, CT, USA\\
\normalsize $^{6}$ Department of Applied Mathematics \& Statistics, Stony Brook University, Stony Brook, NY, USA\\
\normalsize $^{7}$ Department of Computer Science, Stony Brook University, Stony Brook, NY, USA\\
\\
\normalsize  $^{*}$ These authors contributed equally to this work.\\
\normalsize  $^{\dagger}$ Corresponding author(s). E-mail(s): \href{mailto:chenyu.you@yale.edu}{chenyu.you@yale.edu}\\
\normalsize  Contributing authors: \href{mailto:haocheng.dai@utah.edu}{haocheng.dai@utah.edu}; \href{mailto:yifei.min@yale.edu}{yifei.min@yale.edu}; \\
\normalsize  \href{mailto:jasjeet.sekhon@yale.edu}{jasjeet.sekhon@yale.edu}; \href{mailto:sarang.joshi@utah.edu}{sarang.joshi@utah.edu}; \href{mailto:james.duncan@yale.edu}{james.duncan@yale.edu}
\vspace{0.5ex}
}

\date{}
\begin{document}

\maketitle

\begin{abstract}
\noindent
Machine learning models often rely on simple spurious features -- patterns in training data that correlate with targets but are not causally related to them, like image backgrounds in foreground classification. 
This reliance typically leads to imbalanced test performance across minority and majority groups. 
In this work, we take a closer look at the fundamental cause of such imbalanced performance through the lens of memorization, which refers to the ability to predict accurately on \textit{atypical} examples (minority groups) in the training set but failing in achieving the same accuracy in the testing set.
This paper systematically shows the ubiquitous existence of spurious features in a small set of neurons within the network, providing the first-ever evidence that memorization may contribute to imbalanced group performance.
Through three experimental sources of converging empirical evidence, we find the property of a small subset of neurons or channels in memorizing minority group information.
{Inspired by these findings, we hypothesize that spurious memorization, concentrated within a small subset of neurons, plays a key role in driving imbalanced group performance.}
To further substantiate this hypothesis, we show that eliminating these unnecessary spurious memorization patterns via a novel framework during training can significantly affect the model performance on minority groups. 
Our experimental results across various architectures and benchmarks offer new insights on how neural networks encode core and spurious knowledge, laying the groundwork for future research in demystifying robustness to spurious correlation. Our codes are available in {{\href{https://github.com/aarentai/Silent-Majority}{here}}}.
\end{abstract}
\vspace{5ex}

\section{Introduction}
\label{section:intro}

\begin{wrapfigure}{r}{0.5\textwidth}
% \vspace{-20pt}
\includegraphics[width=0.23\columnwidth]{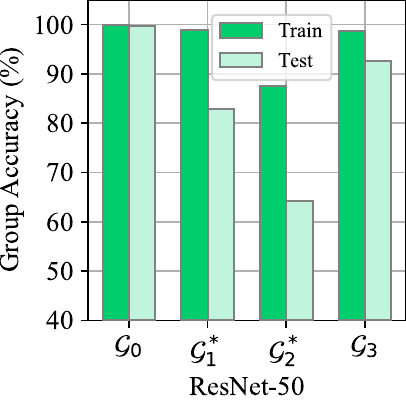}
\includegraphics[width=0.212\columnwidth]{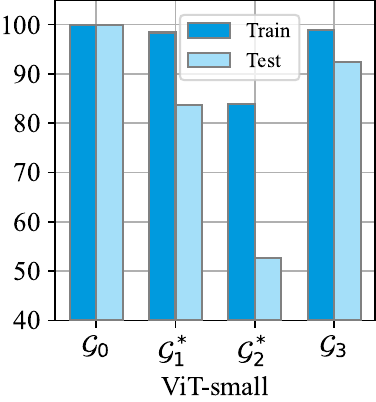}
% \vspace{-10pt}
\caption{\textbf{Imbalanced Group Performance} on Waterbirds. Majority groups ($\mathcal{G}_0$ and $\mathcal{G}_3$) show the minimal gap between training and test accuracy, while minority groups ($\mathcal{G}_1$ and $\mathcal{G}_2$) yield a significantly larger discrepancy. Both models are trained with ERM. 
Here the star superscript (*) in the figure is used to emphasize the minority groups $\mathcal{G}_1$ and $\mathcal{G}_2$.}
\label{fig:train test gap}
\end{wrapfigure}

Machine learning models often achieve high overall performance, yet struggle in minority groups due to \textit{spurious correlations} -- patterns that align with the class label in training data but have no causal relationship with the target \cite{sagawa2019distributionally,geirhos2020shortcut}.
For example, considering the task of distinguishing cows from camels in natural images, it is common to find 95\% cow images with grass backgrounds and 95\% of camel images on sand. 
Models trained using standard Empirical Risk Minimization (ERM) often focus on minimizing the average training error by depending on spurious background attributes (``grass'' or ``sand'') instead of the core characteristics (``cow'' or ``camel''). 
In such settings, models may yield good average accuracy but lead to high error rates in minority groups (``cows on sand'' or ``camel on grass'') \cite{ribeiro2016should,beery2018recognition}. 
This illustrates a fundamental issue: even well-trained models can develop systematic biases from these spurious attributes in their data, thus leading to \textit{alarmingly consistent performance drop} for minority groups where the spurious correlation does not hold.
{Indeed, in Figure~\ref{fig:train test gap}, we present both the training and test accuracy on the majority and minority groups of the Waterbirds benchmark for two popular models: ResNet-50~\citep{he2016deep} and ViT-small~\citep{dosovitskiy2020image}. 
It is clear from Figure~\ref{fig:train test gap} that the test performance is poor on minority groups (1 and 2).}
Moreover, we observe that majority groups have a smaller gap between the training and testing accuracy, as compared to minority groups that have a more significant gap. 
Thus, understanding the underlying causes of this unbalanced performance between the majority and minority groups is crucial to their reliable and safe deployment in various real-world scenarios~\cite{blodgett2016demographic,buolamwini2018gender,hashimoto2018fairness}.

The minority groups are \textit{atypical} examples to neural networks (NNs), as these small subsets of examples bear a similarity to majority groups due to the same spurious attribute, but have distinct labels.
Recent efforts have shown that NNs often `memorize' \textit{atypical} examples, primarily in the final few layers of the model~\cite{baldock2021deep,stephenson2021geometry}, and possibly even in specific locations of the model~\cite{maini2023can}. 
Memorization, in this context, is defined as the neural network's ability to accurately predict outcomes for atypical examples (\eg, mislabeled examples) in the training set through ERM training.  
This is in striking analogy to the spurious correlation issue, because 1) the minority examples are atypical examples by definition, and 2) the minority examples are often more accurately predicted during training but poorly predicted during testing (as demonstrated in Figure~\ref{fig:train test gap}). 
Therefore, a natural open question arises: \textit{Does memorization play a role in spurious correlations?}

In this work, we present the first study to systematically understand the interplay of memorization and spurious correlations in deep overparametrized networks.
We undertake our exploration through the following avenues:
1) What makes the comprehensive condition for the existence or non-existence of spurious correlations within NNs? 2) How do NNs handle {atypical} examples, often seen in minority groups, as opposed to {typical} examples from majority groups? and 3) Can NNs differentiate between these {atypical} and {typical} examples in their learning dynamics?

% \blue{
To achieve these goals, we show the existence of a phenomenon named \textbf{spurious memorization}.
We define `spurious memorization' as the ability of NNs to accurately predict outcomes for atypical (\ie, minority) examples during training by deliberately memorizing them in certain part of the model.
Indeed, we first identify that a small set of neurons is critical for memorizing minority examples. 
These critical neurons significantly affect the model performance on minority examples during training, but only have minimal influence on majority examples. 
Furthermore, we show that these critical neurons only account for a very small portion of the model parameters. 
Such a memorization by a small portion of neurons causes the model performance on minority examples to be non-robust, which leads to the poor testing accuracy on minority examples despite the high training accuracy. 
Overall, our study offers a potential explanation for the differing performance patterns of NNs when handling majority and minority examples.
% }

\begin{wrapfigure}{r}{0.5\textwidth}
\includegraphics[width=0.245\columnwidth]{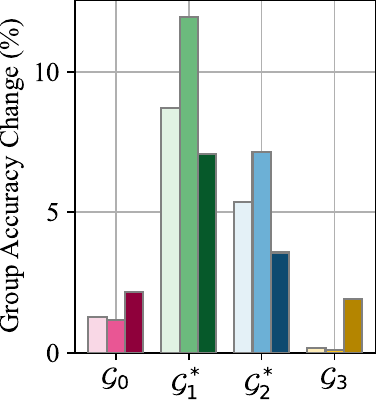}
\includegraphics[width=0.245\columnwidth]{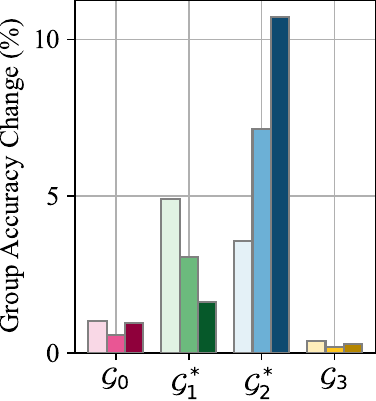}
\caption{\textbf{Group accuracy change by pruning Top-$k$ neuron(s) with gradient-based (left) and magnitude-based (right) criterion.} Within each group $\mathcal{G}$, three bars with gradated hues indicate the accuracy shift after zeroing out the neurons with the top-1, top-2, and top-3 largest gradients or magnitudes, respectively. 
Note that the minority groups $\mathcal{G}_1$ and $\mathcal{G}_2$ are emphasized with the star superscript (*).
}
\label{fig:unstructured zero out barplot}
\end{wrapfigure}
Our systematic study is performed in two stages.
In Stage I, to verify the existence of critical neurons, we identify two experimental sources to trace spurious memorization at the neuron and layer level.
These two sources are \textit{unstructured tracing} (assessing the role of neurons within the entire model for spurious memorization using heuristics including weight magnitude and gradient) and
\textit{structured tracing} (assessing the role of neurons within each individual layer with similar heuristics). 
Specifically, by evaluating the impact of spurious memorization via unstructured and structured tracing at the magnitude and gradient level (Section~\ref{sec:stage 1}), we observe a substantial decrease in minority group accuracy, contrasting with a minimal effect on the majority group accuracy.
This suggests that at unstructured and structured level, the learning of minority group opposes the learning of majority group, and indicates that 
1) critical neurons for spurious memorization indeed exist within NNs; 2) both gradient and magnitude criteria are effective tools for identifying these critical neurons; and 3) NNs tend to memorize typical examples from majority groups on a global scale, whereas {a miniature set of nodes (\ie~critical neurons) is involved in the memorization of minority examples to a greater extent than other neurons.}
Overall, we provide converging empirical evidence to confirm the existence of critical neurons for spurious memorization.

In Stage~II, inspired by the observations above, we develop a framework to investigate and understand the essential role of critical neurons in spurious memorization that would incur the imbalanced group performance of NNs.
Specifically, we construct an auxiliary model which is an adaptively pruned version of the target model, and then contrast the features of this auxiliary model with those of the target model. 
Our motivation comes from recent empirical finding \cite{hooker2019compressed} that pruning can improve a network's robustness to accurately predict rare and atypical examples (minority groups in our case). 
This allows the target model to identify and adapt to various spurious memorization at different stages of training, thereby progressively learning more balanced representations across different groups. 
Through extensive experiments with our training algorithm across a diverse range of architecture, model sizes, and benchmarks, we confirm that the critical neurons have emergent spurious memorization properties, thereby more friendly to pruning. 
More importantly, we show that majority examples, being memorized by the entire network, often yield robust test performance, whereas minority examples, memorized by a limited set of critical neurons, show poor test performance due to the miniature subset of neurons. 
This provides a convincing explanation for the imbalanced group performance observed in the presence of spurious correlations.

Concretely, we summarize our contributions as follows: (1) To the best of our knowledge, we present the first systematic study on the role of different neurons in memorizing different group information, and confirm the existence of critical neurons where memorization of spurious correlations occurs. (2) We show that modifications to specific critical neurons can significantly affect model performance on the minority groups, while having almost negligible impact on the majority groups. (3) We propose \textit{spurious memorization} as a new perspective on explaining the behavior of critical neurons in causing imbalanced group performance between majority and minority groups.

\section{Results}
\subsection{Identifying the Existence of Critical Neurons}
\label{sec:stage 1}

In this section, we validate the existence of \textit{critical neurons} in the presence of spurious correlations. We comprehensively examine the underlying behavior of `critical neurons' on the Waterbirds dataset with the ResNet-50 backbone. 
Within this section, the term `neurons' specifically refers to channels in a convolutional kernel. 
It is worth noting that the Waterbirds dataset comprises two majority groups and two minority groups.
For clarity in our discussions and figures, we use the following notations, aligned with the dataset's default setting:
The majority groups are $\mathcal{G}_0$ (Landbird on Land) and $\mathcal{G}_3$ (Waterbird on Water), while the minority groups are $\mathcal{G}_1$ (Landbird on Water), $\mathcal{G}_2$ (Waterbird on Land).

\myparagraph{Notations.} 
In the following discussion, we consider the model as \( f(\boldsymbol{\theta}, \cdot) \), with $\btheta$ representing the collection of all neurons. 
Individual neurons are denoted as $\mathbf{z}_i$, for $i\in[M]\vcentcolon=\{1,\cdots,M\}$, and \( \boldsymbol{\theta} \) can be expressed as \( \boldsymbol{\theta} = \{\mathbf{z}_1, \mathbf{z}_2, \cdots, \mathbf{z}_M\} \). 
For the training data, we use \( \mathcal{D}_0 \), \( \mathcal{D}_1 \), \( \mathcal{D}_2 \), \( \mathcal{D}_3 \) to represent the datasets, where \( \mathcal{D}_j \) comprises examples from group \( \mathcal{G}_j \), for each $j \in\{ 0, 1, 2, 3\} $, respectively. 
Finally, let \( \mathcal{L}_\ce \) signify the cross-entropy loss.
We emphasize that all the group accuracy evaluated before and after pruning in this section is evaluated on the training set, which strictly complies with the definition of memorization from Section~\ref{section:intro}. 
% \blue{}

\subsubsection{Unstructured Tracing} 
\label{sec:unstructured}

To begin with, we adopt unstructured tracing to assess the effect of neurons on spurious memorization across the entire model, using weight magnitude and gradient as criteria. 

For the gradient-based criterion, we begin with a model trained by ERM. We then select the neurons with the largest gradient, measured in the $\ell_2$ norm, across the entire model. 
Zeroing out these neurons, we can then observe the resultant impact on group accuracy. 
To be specific, we compute the loss gradient for each of the 4 Waterbirds groups.
The loss gradient $\mathbf{v}(\cdot)$ on group $j$ w.r.t. neuron $i$ is defined as
\begin{align*}
    \mathbf{v}(i,j) =  \frac{\partial \mathcal{L}_{\ce}(f(\btheta, \mathcal{D}_j))}{\partial \mathbf{z}_i}, i \in\{ 1,\cdots, M\};  j \in\{ 0,\cdots,3\}.
\end{align*}
For each group $j$, we select those neurons $i's$ of which the $\|\mathbf{v}(i,j)\|_2$ are the top-$k$ largest among all $M$ neurons.\footnote{In our experiments, we evaluate cases with $k=1, 2, 3$. We demonstrate that even just pruning the top-1 largest gradient neuron can significantly affect the minority group training accuracy.} We denote the indices of these neurons as $\mathcal{I}_j$, where $\mathcal{I}_j$ is a subset of $\{ 1,\cdots, M\}$.

\begin{figure}[t]
\includegraphics[width=0.48\textwidth]{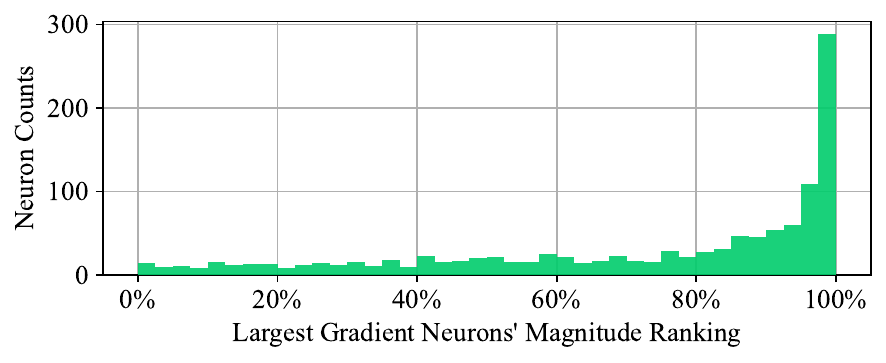}
\includegraphics[width=0.48\textwidth]{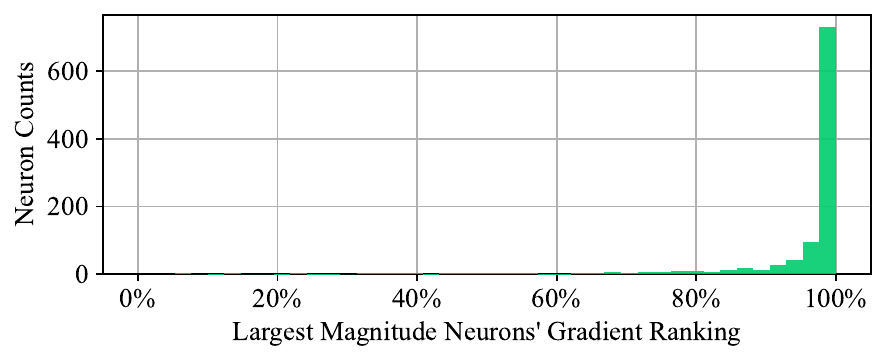}
\caption{\textbf{Largest gradient neurons' magnitude ranking distribution (left) and largest magnitude neurons' gradient ranking distribution (right).} The x-axis percentage refers to the relative ranking of neurons' gradients or magnitudes across the entire network (0\%: smallest, 100\%: largest). 
}
\label{fig:distribution}
\end{figure}

To assess the importance of these selected neurons in memorizing examples, we zero them out and calculate the change in group accuracy on the training set. The pruned model is identified as $f(\mathbf{m}_j \odot \btheta, \cdot)$, where $\mathbf{m}_j$ is a mask with neurons in $\mathcal{I}_j$ being masked. The change in accuracy $\Delta_{\acc}$ for each group $j$ is given by $\Delta_{\acc} (j) = |\acc(\mathcal{D}_j, f(\btheta, \cdot ))  - \acc(\mathcal{D}_j, f(\mathbf{m}_j\odot\btheta, \cdot ))|$, where $\acc$ represents the accuracy.
In our experiments below, all the group accuracy change is based on the following training accuracy: 97.34\% ($\mathcal{G}_0$), 47.83\% ($\mathcal{G}_1$), 69.64\% ($\mathcal{G}_2$), 97.63\% ($\mathcal{G}_3$) \footnote{{For completeness, we also report the baseline test accuracy: 96.98\% ($\mathcal{G}_0$), 35.68\% ($\mathcal{G}_1$), 56.98\% ($\mathcal{G}_2$), and 96.26\% ($\mathcal{G}_3$). The baseline test accuracy follows the same pattern as the training accuracy.}}.

Similarly, when using magnitude as the selection criterion, the tracing procedure remains the same except that we zero-out the neurons with the largest magnitude measured in $\ell_2$ norm. 
That is, instead of $\|\mathbf{v}(i,j)\|_2$, we select neurons with largest $\|\mathbf{z}_i\|_2$.
It is worth noting that the magnitude-based selection approach here is group invariant --- the magnitude used for selection does not vary with the model's input.

\myparagraph{Ablation on the Number of Pruned Neurons.}
We demonstrate that zeroing out the top-1 to top-3 critical neurons can significantly impact the training accuracy of minority groups. 
A natural inquiry arises: are three neurons sufficient? 
In essence, we investigate whether pruning additional neurons can amplify the performance drop. 
Thus, we conduct an ablation study varying the number of pruned neurons. 
The findings are summarized in Table~\ref{table:why top 3} (in Supplementary materials). 
We assert that similar trends persist as observed in Figure~\ref{fig:unstructured zero out barplot}, despite altering the number of pruned neurons. 
Notably, the decline in performance among minority groups ($\mathcal{G}_1$ and $\mathcal{G}_2$) exceeds that of majority groups ($\mathcal{G}_0$ and $\mathcal{G}_3$), even with an increase to 10 neurons pruned.

\myparagraph{Observation and Analysis.} 
In our study, we plot the change in accuracy, $\Delta_{\acc} (j)$, for each group $j$ as shown in Figure~\ref{fig:unstructured zero out barplot}.
For every group, we consider three scenarios: pruning the top-1, top-2, and top-3 neurons, which corresponds to the 3 bars for each group in Figure~\ref{fig:unstructured zero out barplot} Note that we limit our reporting to the results involving up to 3 critical neurons based on experimental findings indicating that pruning the top-3 neurons is adequate. This decision is supported by the number of pruned neurons, detailed in Supplementary Materials Table~\ref{table:why top 3} (in Supplementary materials).
It can be clearly observed that the accuracy of minority groups exhibits significant shifts, while the accuracy of majority groups shows only minimal impact.
Specifically, for the majority groups $\mathcal{G}_0$ and $\mathcal{G}_3$, the maximum of the group accuracy shifts stands at $2.15\%$ when we zero out the top 3 neurons with the largest gradient. While for minority group $\mathcal{G}_1$ and $\mathcal{G}_2$, the maximum of the group accuracy shifts stands at $11.96\%$ when we zero out the top 2 neurons with the largest gradient. 
This is a sharp contrast between the groups, where accuracy shifts significantly, underscoring the critical role of selected neurons in memorizing minority examples at both gradient and magnitude levels. 
Meanwhile, the substantial contrast in accuracy shifts between majority and minority groups provides initial evidence that the model's performance on minority groups can be solely dependent on a few neurons, occasionally even as few as three or fewer.

\myparagraph{Both the gradient-based and magnitude-based criteria work.} 
Interestingly, we observe that both the gradient-based and magnitude-based criteria can yield similar effects. 
We show in the following that it is attributed to an overlap in the distribution of critical neurons identified by each criterion. 
To delve deeper, in Figure~\ref{fig:distribution}, we analyze the relative magnitude ranking among all neurons for the neurons with the largest gradient, and the relative gradient ranking for neurons with the largest magnitude. 
In the left of Figure~\ref{fig:distribution}, we show the magnitude ranking for the neurons with top 0.01\% largest gradient, and Figure~\ref{fig:distribution} right subfigure demonstrates the gradient ranking for the top 0.01\% largest magnitude neurons. 
In both histograms, there is a noticeable clustering in the rightmost two bins (ranging from 95\% to 100\%). This suggests that the neurons with the highest magnitudes tend to exhibit large gradients, and the neuron with the largest gradient often coincides with a high weight magnitude. 
This finding provides tantalizing evidence of the similar distribution of critical neurons under both criteria and explains the matching phenomenon observed between the two criteria.

\myparagraph{Random Noise and Random Initialization.}
Our experiments thus far offer preliminary evidence for the existence of critical neurons. 
To gain a more comprehensive understanding, we explore alternatives to pruning, especially studying the effects of random initialization and random noise.
These two experiments are motivated by our desire to investigate the effects of perturbation from two perspectives: perturbation on the original neuron weights and perturbation on the pruned neurons. 
By examining these perturbations, we draw more credible supporting evidence on the existence of critical neurons by evaluating the sensitivity of group accuracy to specific neurons more comprehensively.

\begin{wrapfigure}{r}{0.5\textwidth}
\centerline{
\includegraphics[width=0.225\columnwidth]{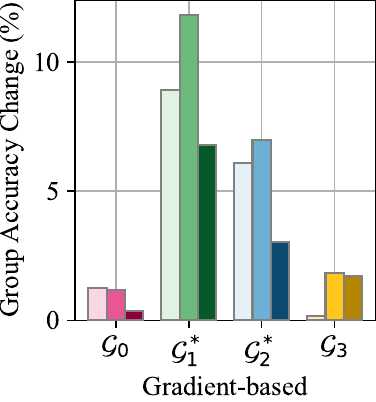}
\includegraphics[width=0.225\columnwidth]{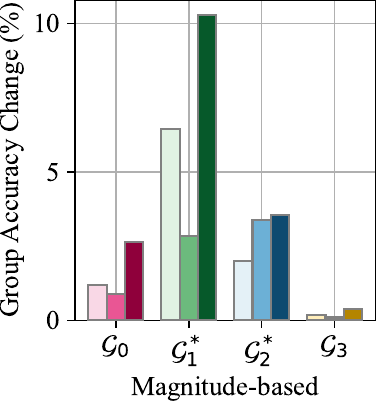}
}
\caption{\textbf{Group accuracy change by random initialize Top-$k$ neuron(s) with gradient-based (left) and magnitude-based (right) criterion.} For each group, 3 bars with gradated hues indicate the accuracy shift after random initializing top-1, top-2, and top-3 neurons with the largest gradient or magnitude, respectively.}
\label{fig:random initialization barplot}
\end{wrapfigure}

$\rhd$ \textbf{How to implement random initialization?} Instead of performing pruning on the selected neurons, we opt to initialize them randomly using a zero-mean Gaussian random variable. That is, we replace the neuron weight $\mathbf{z}_i$ with $\tilde{\mathbf{z}}_i = \epsilon_i$ where $\epsilon_i\sim \mathcal{N}(\mathbf{0}, \bsigma^2)$. The accuracy change is formulated as:
\begin{align*}
    \Delta_{\acc} (j) = |\acc(\mathcal{D}_j, f(\btheta, \cdot ))  - \acc(\mathcal{D}_j, f(\tilde\btheta, \cdot ))|, 
\end{align*} 
where $\tilde\btheta = \{\mathbf{z}_i\}_{i\notin \mathcal{I}_j} \cup \{\tilde{\mathbf{z}}_i\}_{i\in \mathcal{I}_j}$. 
The result is in {Figure}~\ref{fig:random initialization barplot}. 

\begin{wrapfigure}{r}{0.5\textwidth}
\centerline{
\includegraphics[width=0.225\columnwidth]{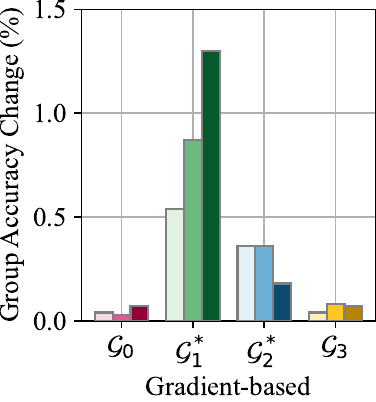}
\includegraphics[width=0.225\columnwidth]{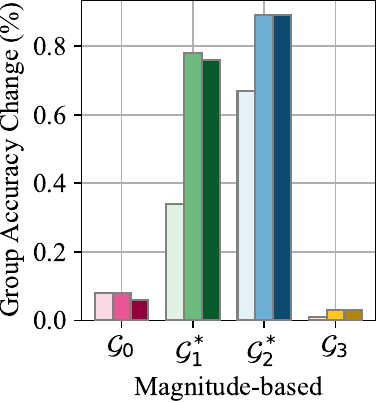}
}
\caption{\textbf{Group accuracy change by adding random noise to Top-$k$ neuron(s) with gradient-based (left) and magnitude-based (right) criterion.} For each group, 3 bars of gradated hues indicate the group accuracy change after adding random noise to top-1, top-2, and top-3 neurons with largest gradient or magnitude.}
\label{fig:random noise barplot}
\vspace{-10pt}
\end{wrapfigure}

$\rhd$ \textbf{How to implement random noise?} We add an extra noise term, which is a zero-mean Gaussian random variable, to the selected neurons, \ie, $\tilde{\mathbf{z}}_i = \mathbf{z}_i + \epsilon_i$, where $\epsilon_i\sim \mathcal{N}(\mathbf{0}, \mathbf{\sigma}^2)$. 
The result is shown in {Figure}~\ref{fig:random noise barplot}. 

In Figure~\ref{fig:random initialization barplot} and ~\ref{fig:random noise barplot}, we found that 1) the results from random initialization ($\mathbf{\sigma}=0.005$) closely resemble those from the pruning method. 
Notably, the minority groups show a $2\% - 12\%$ shift in group accuracy compared to the majority groups' $0\% - 2.6\%$ shift, since random initialization converges to pruning as the standard deviation of the Gaussian random variables $\{\epsilon_i\}_{i\in\mathcal{I}_j}$ decreases;
2) with random noise added ($\mathbf{\sigma}=0.005$), the accuracy changes in minority groups still surpass those in majority groups. We also observe that the extent of accuracy change with random noise is much smaller than that observed with random initialization and pruning. This occurs because, in the presence of random noise, neuron values are not reset to a mean-zero state, allowing their initial values to impact the model's performance across different groups. 
{Moreover, we conduct the additional unstructured tracing experiment on CelebA, as shown in Table \ref{table:celebA unstructured train} (in Supplementary materials). We can see that the results for the minority group of CelebA align with those obtained on the Waterbirds dataset.
As shown in Table \ref{table:compare train test waterbirds}, our results in the Waterbirds dataset reveal that when modifying critical neurons, the training accuracy for minority groups ($\mathcal{G}_1$ and $\mathcal{G}_2$) consistently drops across nearly all experimental setups, whereas the corresponding test accuracy remains relatively stable. 
This discrepancy reinforces the interpretation that these neurons play a memorization role.
Furthermore, the raw values reported in Table \ref{table:raw change} further substantiate these findings, showing that for minority groups, training accuracy decreases consistently in all experimental settings and for every choice of $k$ in the top-$k$ analysis. Together, these results provide compelling evidence that identified critical neurons are primarily responsible for memorization in minority groups, rather than affecting overall generalization.} 
Additional experimental results on random initialization and random noise with various standard deviations can be found below.

\myparagraph{Random Initialization and Random Noise.}
For all random initialization (Figure~\ref{fig:appendix_random_init} in Supplementary materials) and noise-adding (Figure~\ref{fig:appendix_random_noise} in Supplementary materials) experiments in Section~\ref{sec:stage 1}, we choose the random variable with multiple standard deviations to validate the existence of critical neuron. For each subfigure in Figure~\ref{fig:appendix_random_init} and Figure~\ref{fig:appendix_random_noise} (in Supplementary materials), the results is averaged over 10 random seeds.

Overall, regardless of the scale variation in the accuracy shifts, our experiments using pruning, random initialization, and random noise consistently demonstrate that the accuracy of minority groups is significantly sensitive to the alteration of a handful of selected neurons. {This finding suggests that a small subset of \textit{critical neurons} contributes more significantly to the memorization of minority examples during training than other neurons.}
Moreover, it validates that both gradient-based and magnitude-based criteria are effective in identifying these critical neurons.

\subsubsection{Structured Tracing}

In unstructured tracing (Section~\ref{sec:unstructured}), we select neurons from the entire model without considering any sub-structures (\ie, layers, blocks) of networks. 
To gain a deeper understanding of how these sub-structures influence memorization, we use structured tracing for probing and comprehending the role of sub-structures in the networks.

Specifically, we begin by fixing a particular layer, and then selecting neurons within the layer to assess the importance of these neurons in memorizing examples from groups. 
We still employ either gradient-based or magnitude-based criterion for neuron selection, but the scope of this specific experiment is confined to the individual layer. 
This process is identically repeated for each layer in the entire model.

\myparagraph{Observation and Analysis.} 
In Figure~\ref{fig:structured heatmap} (in Supplementary materials), we employ a heatmap to visualize how accuracy changes across different groups when we selectively zero-out a subset of neurons within a specific layer. 
What becomes evident is that deactivating the same number of neurons with the highest gradients or magnitudes within a layer consistently leads to a more significant shift in accuracy for the minority groups compared to the majority groups.
This difference is clearly discernible in the brighter color associated with the minority groups $\mathcal{G}_1$ and $\mathcal{G}_2$ in the middle two rows.
Furthermore, we notice that these within-layer critical neurons appear to be distributed across multiple layers in the early stages of the model, rather than being confined to the final few layers.
This finding aligns with the literature which indicates that the memorization of atypical examples can be distributed and localized throughout the neural networks~\citep{maini2023can}.

\subsection{Spurious Memorization by Critical Neurons}
\label{sec:stage 2}

{In Section~\ref{sec:stage 1}, our experiments have empirically demonstrated the presence of a small set of critical neurons involved in the memorization of minority examples during training.}
This underscores the role of spurious memorization as a significant factor in imbalanced group performance. 
In this section, we take a further step in demystifying the cause of imbalanced group performance under spurious correlation, particularly focusing on the discrepancy in the test accuracy between majority and minority groups.

To further validate the hypothesis that spurious memorization is a key factor in the imbalanced group performance, we investigate whether countering spurious memorization during training could lead to improved test accuracy on minority groups. 
Our findings affirmatively answer this question. 
By specifically targeting and removing spurious memorization via a specialized fine-tuning framework, we observe a consistent improvement in the test accuracy for minority groups. We report extensive experimental results across different model architectures, including ResNet-50 and ViT-Small, and on benchmark datasets including Waterbirds~\citep{sagawa2019distributionally,wah2011caltech} and CelebA~\citep{liu2015deep}, providing comprehensive analysis on the effects of spurious memorization on imbalanced group performance.

\subsubsection{Interference with Spurious Memorization}
\label{sec:stage2 method}

\myparagraph{Our Framework.}
Figure~\ref{fig:model} (in Supplementary materials) summarizes our fine-tuning framework for analyzing spurious memorization. 
By default, our framework is built upon simCLR~\citep{chen2020simple}, adhering to its key components such as data augmentations and the non-linear projection head. The primary distinction between ours and simCLR is centered around two models: a target model and an auxiliary model. The auxiliary model is essentially a pruned version of the target model, where certain critical neurons are masked while the remaining neurons retain the same weights as the target model. 
This allows the framework to feed two augmented images into separate models, yielding two distinct feature representations for contrasting with each other.

More specifically, we begin with the target model, represented as $f(\btheta, \cdot)$, where $\btheta$ denotes the model weights. These weights are initialized by pretraining the model using ERM. The next stage involves fine-tuning the target model. To this end, we construct a pruned model, $f(\mathbf{m} \odot \btheta, \cdot)$, with $\mathbf{m}$ being a masking vector. The mask is created based on criteria derived from either gradient or magnitude, as inspired in Section~\ref{sec:stage 1}. In our experiments, we zero-out the top 0.01\% of neurons based on their $\ell_2$-norm of their gradient or magnitude, where 0.01\% serves as a hyperparameter.
 % until convergence\footnote{Empirically, we observe convergence after $\sim$40 epochs.}

$\rhd$ \textbf{How is the gradient calculated?} It is worth noting that for gradient calculation, we do not rely on group labels as in Section~\ref{sec:stage 1}, but instead use the model's predictions as pseudo labels for sample selection. During each epoch, we calculate the cross-entropy loss for each sample, select the top 256 samples with the highest loss, and randomly sample 128 out of them to form the batch for gradient computation. 

During training, contrasting output features of the two models -- target and auxiliary -- enables adaptive online identification of critical neurons for the current target model. 
This approach implicitly gives greater emphasis to these samples in the loss function, effectively tailoring the training process to bolster the model in recalling these challenging forgotten examples. 
In particular, the \textbf{key innovation} in our training framework is to restrict the target model's tendency to memorize atypical examples using only a small set of neurons.
Inspired by the observation \cite{hooker2019compressed} that pruning reduces a model's prediction accuracy on rare and atypical instances, we enforce feature alignment by adopting the NT-Xent loss ~\citep{sohn2016improved} as we find that these samples typically exhibit the greatest prediction disparities between the pruned and non-pruned models. 
Utilizing pruning as an experimental tool will amplify the prediction disparity between the pruned and non-pruned models, resulting in an implicit rebalancing of the loss.

Consider an arbitrary input image $\mathbf{x}$, and denote by $\mathbf{x}'$ and $\mathbf{x}''$ two augmentations of $\mathbf{x}$.
The loss term on the input $\mathbf{x}$ (together with its positive and negative pairs) 
can be formulated as:
\begin{align}\label{eq:loss ntxent}
    &\mathcal{L}_{\textnormal{NT-Xent}}(\btheta, \mathbf{x})
    = -\log \frac{\exp(\textnormal{sim}(\mathbf{r},\mathbf{r}_p)/\tau)}{\sum_k \exp(\textnormal{sim}(\mathbf{r},\mathbf{r}_k)/\tau)},
\end{align} where
$\mathbf{r}$ is the output feature of input $\mathbf{x}$ by the target model $f(\btheta, \mathbf{x}')$,  $\mathbf{r}_p$ is the output feature of the auxiliary model $f(\mathbf{m}\odot\btheta, \mathbf{x}'')$, and $\mathbf{r}_k$ is the output feature of negative pairs. 
And $\tau$ is the loss temperature, 
and $\textnormal{sim}(\cdot,\cdot)$ is the cosine similarity: $\textnormal{sim}(\mathbf{u}, \mathbf{v}) = \mathbf{u}\cdot\mathbf{v} / (\|\mathbf{u}\|\cdot \|\mathbf{b}\|)$. 

Additionally, we incorporate a supervised loss for the target model. Interestingly, we found that the Mean Squared Error (MSE) loss had a more pronounced effect than the Cross Entropy (CE) loss in our experiments \citep{hui2021evaluation}. Thus, we adopt the MSE loss, as defined as:
\begin{align}\label{eq:loss mse}
    \mathcal{L}_{\textnormal{MSE}} (\btheta, \mathbf{x}, \mathbf{y}) = \|\hat{\mathbf{y}} - \mathbf{y}\|_2^2,
\end{align} where $\mathbf{y}$ is the one-hot vector of the ground-truth class of input $\mathbf{x}$, and $\hat{\mathbf{y}}$ is the model prediction vector. 
The final loss function is formulated as:
\begin{align}\label{eq:total loss}
    \mathcal{L}_\textnormal{total} (\btheta, \mathbf{x}, \mathbf{y}) = \mathcal{L}_{\textnormal{NT}}(\btheta, \mathbf{x}) + \lambda \mathcal{L}_{\textnormal{MSE}} (\btheta, \mathbf{x}, \mathbf{y}),
\end{align}
where $\lambda>0$ is a hyperparameter for balancing loss terms.

In summary, we initiate the process by pretraining the target model using ERM and subsequently fine-tune it further using the above-mentioned framework for a few additional epochs. 
At the start of each fine-tuning epoch, we create an auxiliary model by pruning a small portion of neurons from the target model based on either gradient or magnitude. 
The target and auxiliary models are then trained to align their output features with each other, fostering robust learning.

\subsubsection{Removing Spurious Memorization Improves Group Robustness}
\label{sec:stage2 result}
\begin{wrapfigure}{r}{0.5\textwidth}
\centerline{
\includegraphics[width=0.24\columnwidth]{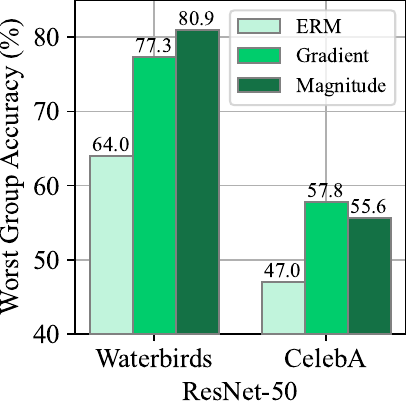}
\includegraphics[width=0.218\columnwidth]{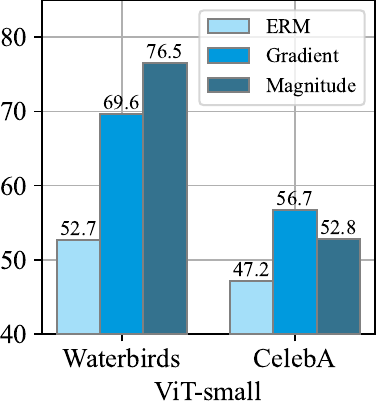}
}
\caption{\textbf{Comparison of Worst Group Accuracy (WGA) across various architectures and datasets.} The light-colored bars corresponds to the standard ERM training, and the dark color corresponds to our proposed framework.}
\label{fig:wga comparison}
\end{wrapfigure}

In this study, our primary objective is to investigate whether mitigating spurious memorization can lead to an enhancement in the test accuracy of minority groups.
The findings are illustrated in Figure~\ref{fig:wga comparison}, where we compare the Worst Group Accuracy (WGA) between standard ERM training and our proposed framework. The WGA for ERM training is evaluated by the testing set with the best-performing checkpoint (on the validation set) among the first 100 epochs. Notably, we observe a significant increase in WGA under all scenarios. Specifically, with the {ResNet-50} backbone, we observe a significant 16.87\% and 10.77\% improvement in WGA for the Waterbirds and CelebA datasets, respectively. Similarly, on {ViT-Small} model, we observe 23.83\% and 9.45\% improvements in WGA. {Furthermore, the statistical summary is included in Table \ref{table:summary statistics} (in Supplementary materials).}

It is important to highlight that our auxiliary model is essentially a pruned version of the target model, with only 0.01\% of the neurons being masked. Despite this seemingly small modification, the consistent performance boost in WGA across different architectures and datasets is strikingly remarkable. This improvement suggests that by strategically disrupting the spurious memorization mechanism through contrasting two model branches, we can guide the target model to learn atypical or minority examples more robustly.
These findings lend further support to our hypothesis that spurious memorization contributes to imbalanced group performance. 
Additionally, we have conducted comprehensive ablation studies to explore various aspects of our framework, including the choice of kick-in and fine-tuning epoch, loss balancing term $\lambda$, pruning ratio, different gradient sources, and loss function. 

\subsubsection{Ablation on Hyper-parameters}
% \label{sec:ablation}

In this subsection, we perform extensive ablation studies to offer a more comprehensive perspective of the framework. 
All the ablation experiments are conducted using the Waterbirds dataset with the ResNet-50 model. 

\myparagraph{Ablation on Loss Functions.}
In Section~\ref{sec:stage2 method} we use MSE as one of the loss terms in the model fine-tuning (see Eqs. \eqref{eq:loss mse} and \eqref{eq:total loss}).  
Here we compare MSE with Cross Entropy (CE) loss. 
Using CE, the final loss becomes $\mathcal{L}_\textnormal{total} (\btheta, \mathbf{x}, \mathbf{y}) = \mathcal{L}_{\textnormal{NT}}(\btheta, \mathbf{x}) + \lambda \mathcal{L}_{\textnormal{CE}} (\btheta, \mathbf{x}, \mathbf{y})$, as compared to Eq. \eqref{eq:total loss}. 
The result is shown in Table~\ref{table:loss function}. 
We observe that MSE is more effective in terms of WGA gain than CE under the same pruning percentage.
Still, both choices manifest significant WGA gain against ERM, corroborating our hypothesis that spurious memorization in the critical neurons might play a critical role in imbalanced group performance. 

\myparagraph{Ablation on Kick-in Epoch.}
In our training framework introduced in Section~\ref{sec:stage2 method}, we first pretrain the target using ERM for 40 epochs and then switch to a fine-tuning stage with loss function Eq. \eqref{eq:total loss}. 
In other words, our framework kicks in at epoch 40. 
Here we test different choices of the kick-in epoch.
The result is shown in Table~\ref{table:kickin epoch} (in Supplementary materials). 
Overall, we see that epoch 30 is not effective, while the choice of 40, 50, and 60 all return meaningful returns. 
This reason is that the ERM training of the target model has not converged yet at epoch 30.

\myparagraph{Ablation on Number of Fine-tuning Epochs.}
We then compare the number of epochs for the fine-tuning stage. 
The result is shown in Table~\ref{table:finetuning epochs} (in Supplementary materials).
Observe that, there is no difference between the result for using 20 or 30 fine-tuning epochs.
This is because the best model check-point appears within 20 epochs. The result indicates that fine-tuning for more epochs is  unnecessary.

\myparagraph{Ablation on Data Source for Gradient Calculation.}
We compare different source of gradient in the calculation of gradient-based criterion. 
As introduced in Section~\ref{sec:stage2 method}, the neuron gradient is computed on a selected subset of training data.
By default, this subset is chosen as the worst predicted examples by the target model in terms of the CE loss. 
From Table~\ref{table:gradient source} (in Supplementary materials), we observe that, calculating the gradient on the subset of worst predicted examples from the minority groups does not show any benefit.
Considering the fact that using minority groups as the gradient source requires access to the group label which is sometimes unavailable, we suggest using the full training set as the gradient source.

\myparagraph{Ablation on Loss Term Ratios.}
We compare different choice of the loss term ratio $\lambda$ in Eq.~\eqref{eq:total loss}. 
The result is shown in Table~\ref{table:loss ratio} (in Supplementary materials).

\myparagraph{Ablation on Pruning Percentage.}
We compare different choice of pruning percentage for both the gradient-based criterion and magnitude-based criterion. 
The result is shown in Table~\ref{table:pruning percentage} (in Supplementary materials).

\myparagraph{Ablation on Combined Pruning Criteria.}
We test a mixed pruning criterion which combines the gradient-based criterion with the magnitude-based one. 
The result is shown in Table~\ref{table:combining criteria} (in Supplementary materials).

\subsubsection{Visualization Results}
\label{sec:other results}

To interpret the outcome of the trained neural networks by ERM and our fine-tuning strategy, we visualize the GradCAM on ResNet-50 trained by solely ERM and our fine-tuning strategy. The target layer is set to  \texttt{layer4.2.conv3.weight}, and the target dimension in output feature is set to dimension 0. Figure~\ref{fig:gradcam} (in Supplementary) clearly shows that by our fine-tuning strategy, the neural network shifts its focus from the spurious element (\ie, background) to the main object (\ie, bird).

\section{Discussion}

In this paper, we conduct the systematic investigation aimed at uncovering the root structural cause of imbalanced group performance in the presence of spurious correlations. 
This phenomenon is characterized by both majority and minority groups achieving high training accuracy, yet minority groups experiencing reduced testing accuracy. 
Our comprehensive study verifies the presence of spurious memorization, a mechanism involving critical neurons significantly influencing the accuracy of minority examples while having minimal impact on majority examples. 
Building upon these key findings, we demonstrate that by intervening with these critical neurons, we can effectively mitigate the influence of spurious memorization and enhance the performance on the worst group. 
Our findings shed light on the reasons behind NNs demonstrating robust performance with majority groups but struggling with minority groups, and establish spurious memorization as a pivotal factor contributing to imbalanced group performance. 
We hope that our discoveries offer valuable insights for practitioners and serve as a foundation for further exploration into the intricacies of memorization in the presence of spurious correlations.

Mitigating spurious correlations in machine learning and statistical models is a key step towards crafting more reliable and trustworthy medical AI. Our research uncovers that by eliminating spurious memorization, we can pinpoint critical neurons, whose modification significantly influences the model's performance, particularly in recognizing minority groups. Concerning privacy risks, these are relatively low in our approach, as the analysis requires existing access to the dataset and the capability to train models. Looking forward, our future research will aim to address challenges within the broader scope of spurious correlations, extending beyond vision applications to include language datasets, among others. This expansion will help in developing AI solutions that are more versatile and universally applicable.

\section{Methods}
\label{section:experiments}

\subsection{Experimental Setup}
\label{sec:general details}

\myparagraph{Datasets and Models.} 
In our study, we conduct experiments on two popular benchmark datasets for spurious correlation: Waterbirds~\citep{sagawa2019distributionally,wah2011caltech}, and CelebA~\citep{liu2015deep}. 
We comprehensively evaluate the extent to which spurious memorization exists in the large pre-trained models ({ResNet-50}~\citep{he2016deep} and {ViT-Small}~\citep{dosovitskiy2020image}) on ImageNet~\citep{deng2009imagenet}. 
Note that we report the average performance of 10 independent runs with different random seeds for experiments including unstructured tracing and structured tracing. 
In the experiments detailed in Section~\ref{sec:stage 2}, we strictly adopt the standard dataset splits for both Waterbirds and CelebA, following the setting in~\citep{idrissi2022simple}.
Our adoption of ResNet or ViT models pre-trained on ImageNet is consistent with the main literature~\citep{kirichenko2022last,qiu2023simple,yang2023mitigating}.
Furthermore, the high baseline accuracy achieved by pre-trained models is critical for studying memorization, which is a focal point of our study.

\myparagraph{Identification of Critical Neurons.}
For identifying critical neurons, we utilize two key metrics: gradient-based and magnitude-based criteria. Here, \textit{gradient} refers to the gradients calculated during backpropagation with respect to a specific data batch. \textit{Magnitude}, on the other hand, is determined by the norm of neuron weights. 
Details of data batch selection are given in Section~\ref{sec:unstructured} and Section~\ref{sec:stage 2}.

\myparagraph{Neurons and Layers.} 
For our study on convolutional neural networks (e.g., using {ResNet-50} as the backbone), we consider channels as the basic units in order to preserve the channel structure, as suggested in prior work~\citep{maini2023can}. On the other hand, for our study involving Vision Transformer (e.g., using {ViT-Small} as the backbone), we consider individual neurons as the basic units. Therefore, for ease of reference in our study, we use the term `neuron' to collectively refer to both \textit{channels} in {ResNet-50} and \textit{neurons} in {ViT-Small}.

\myparagraph{Experimental Setup.} 
In all our experiments, we keep the experimental setup consistent. 
We use a single NVIDIA Titan RTX GPU. 
We conduct our experiments using PyTorch 1.13.1+cu117 and Python 3.10.4, to ensure reproducibility.

\myparagraph{Data Preprocessing.}
Our dataset preprocessing remains consistent across all datasets and experiments. 
For details, please refer to Table~\ref{table:augmentation} (in Supplementary materials). These steps ensure that the resulting image size is $224\times224$ pixels, suitable for both ResNet-50 and ViT-S/16@224px. 
Following these augmentation steps, we normalize the image by subtracting the average pixel values (mean=[0.485, 0.456, 0.406]) and dividing by the standard deviation (std=[0.229, 0.224, 0.225]). This normalization procedure aligns with the approach used in CLIP~\cite{radford2021learning}.
No further data augmentation is applied after these steps.

\myparagraph{Hyperparameters.} 
A comprehensive collections of hyperparameters and their values is presented in Table~\ref{table:hyperparameter} (in Supplementary materials).

\myparagraph{Implementation Details.}
In Section~\ref{sec:stage 1}, we implemented the ERM with specific configurations. We utilized an Adam optimizer with a weight decay of 0 and a momentum of 0.9. The learning rate was set at a fixed value of $1\times10^{-4}$, and the models were trained for a total of 100 epochs. To dynamically adjust the learning rate, we employed a ReduceLROnPlateau scheduler, which reduces the learning rate by a factor of 0.5 after a patience of 3 epochs.
For the ERM method, the batch size used in the ERM process was set to 256, and the model's parameters were initialized with \texttt{torchvision.models.ResNet50\_Weights.IMAGENET1K\_V2}. In our experiments involving gradient-based neuron modification, we first curated a set of 256 worst-performing samples. From this set, we randomly sampled 128 samples for calculating both the loss and the gradient.
Our model selection process remained consistent across all methods. After each epoch, we evaluated the model's performance on the validation set and selected the model with the highest worst-group accuracy as the final model for testing. It's important to note that all accuracy metrics reported in this paper are derived from the test set.

Figure~\ref{fig:distribution} presents an analysis of the distribution and relative rankings of neurons in two aspects: their gradient magnitude for neurons with the highest magnitudes, and their magnitude for neurons with the largest gradients.
Specifically, in a model's convolutional layer, each neuron possesses a magnitude (the norm of the weight) and a gradient magnitude (the norm of the gradient). In the upper part of Figure~\ref{fig:distribution}, we select neurons from the convolutional layer that are in the top 0.01\% in terms of gradient magnitude. We then calculate the percentage rank of these selected neurons based on their magnitude compared to all neurons in the convolutional layer. Similarly, in the lower part of Figure~\ref{fig:distribution}, we select neurons that are in the top 0.01\% in terms of weight magnitude and calculate the percentage rank of these based on their gradient magnitude relative to all neurons in the convolutional layer.
Both histograms in Figure~\ref{fig:distribution} show a significant concentration in the two rightmost bins (which represent the range from 95\% to 100\%). This indicates that neurons with the highest weight magnitudes tend to have large gradients, and neurons with the highest gradients often have substantial weight magnitudes. This observation provides intriguing evidence of a similar distribution pattern for critical neurons under both criteria, explaining the observed correlation between these two metrics.

In Section~\ref{sec:stage 2}, all the methods we assess utilize an Adam optimizer with a weight decay of 0 and a momentum of 0.9. The learning rate is held constant at $2\times10^{-4}$ throughout the training process, which lasts for 20 epochs. Additionally, we implement a ReduceLROnPlateau scheduler, which dynamically adjusts the learning rate. This scheduler reduces the learning rate by a factor of 0.5 and waits for 1 epoch before making adjustments.
In our experiments involving gradient-based pruning, we selected 256 of the poorest-performing samples and then randomly sampled 128 from this group to calculate both the loss and the gradient.
Our model selection standard remains uniform across all methods. After each epoch, we evaluate the model's performance on the validation set and choose the one that achieves the highest worst-group accuracy as the final model for testing. It's important to note that all accuracy metrics presented in this paper are derived from the test set. For selecting positive and negative pairs, within our framework, we define positive pairs in contrastive learning as the output features that originate from the same input image. Conversely, when dealing with output features from different images within the batch, we regard them as negative samples relative to the current feature.

\section*{Data availability}
The Waterbirds dataset is available at \url{https://github.com/kohpangwei/group_DRO}, formed from \url{https://www.vision.caltech.edu/datasets/cub_200_2011/} and \url{http://places2.csail.mit.edu/}. And the CelebA dataset is available at \url{http://mmlab.ie.cuhk.edu.hk/projects/CelebA.html/}. All requests from institution-affiliated researchers for access to processed data for purposes of study validation will be considered and should be directed to C.Y. (\href{mailto:chenyu.you@yale.edu}{chenyu.you@yale.edu}), and will be handled within 1 month.

\section*{Code availability}
{The code that supports the findings of this study is available at \url{https://github.com/aarentai/Silent-Majority}.}

\section*{Acknowledgements} 
C.Y. and J.S.D. were supported by NIH grants R01CA206180 and R01HL121226.
H.D. and S.J. were supported by NSF grant DMS-1912030.
The authors want to thank all the anonymous reviewers and editors for their constructive comments and suggestions that substantially improved this paper. C.Y. is the corresponding author of this paper.

\section*{Authors contribution} 
C.Y., H.D., and Y.M. developed and implemented new machine learning methods, benchmarked machine learning models and analyzed model behavior, all under the guidance of J.S.S., S.J., J.S.D., and performed a validation study to evaluate its effects. All authors discussed the results and contributed to the final manuscript. C.Y., H.D., and Y.M. designed the study.

\section*{Competing interests} 
The authors declare no competing interests.

\clearpage

\bibliographystyle{plainnat}
\bibliography{refs}

\begin{thebibliography}{24}
\providecommand{\natexlab}[1]{#1}
\providecommand{\url}[1]{\texttt{#1}}
\expandafter\ifx\csname urlstyle\endcsname\relax
  \providecommand{\doi}[1]{doi: #1}\else
  \providecommand{\doi}{doi: \begingroup \urlstyle{rm}\Url}\fi

\bibitem[Baldock et~al.(2021)Baldock, Maennel, and Neyshabur]{baldock2021deep}
Robert Baldock, Hartmut Maennel, and Behnam Neyshabur.
\newblock Deep learning through the lens of example difficulty.
\newblock In \emph{Advances in Neural Information Processing Systems},
  volume~34, pages 10876--10889, 2021.

\bibitem[Beery et~al.(2018)Beery, Van~Horn, and Perona]{beery2018recognition}
Sara Beery, Grant Van~Horn, and Pietro Perona.
\newblock Recognition in terra incognita.
\newblock In \emph{Proceedings of the European conference on computer vision
  (ECCV)}, pages 456--473, 2018.

\bibitem[Blodgett et~al.(2016)Blodgett, Green, and
  O’Connor]{blodgett2016demographic}
Su~Lin Blodgett, Lisa Green, and Brendan O’Connor.
\newblock Demographic dialectal variation in social media: A case study of
  african-american english.
\newblock In \emph{Proceedings of the 2016 Conference on Empirical Methods in
  Natural Language Processing}, pages 1119--1130, 2016.

\bibitem[Buolamwini and Gebru(2018)]{buolamwini2018gender}
Joy Buolamwini and Timnit Gebru.
\newblock Gender shades: Intersectional accuracy disparities in commercial
  gender classification.
\newblock In \emph{Conference on fairness, accountability and transparency},
  pages 77--91. PMLR, 2018.

\bibitem[Chen et~al.(2020)Chen, Kornblith, Norouzi, and Hinton]{chen2020simple}
Ting Chen, Simon Kornblith, Mohammad Norouzi, and Geoffrey Hinton.
\newblock A simple framework for contrastive learning of visual
  representations.
\newblock In \emph{International conference on machine learning}, pages
  1597--1607. PMLR, 2020.

\bibitem[Deng et~al.(2009)Deng, Dong, Socher, Li, Li, and
  Fei-Fei]{deng2009imagenet}
Jia Deng, Wei Dong, Richard Socher, Li-Jia Li, Kai Li, and Li~Fei-Fei.
\newblock Imagenet: A large-scale hierarchical image database.
\newblock In \emph{IEEE conference on computer vision and pattern recognition},
  pages 248--255. IEEE, 2009.

\bibitem[Dosovitskiy et~al.(2021)Dosovitskiy, Beyer, Kolesnikov, Weissenborn,
  Zhai, Unterthiner, Dehghani, Minderer, Heigold, Gelly, Uszkoreit, and
  Houlsby]{dosovitskiy2020image}
Alexey Dosovitskiy, Lucas Beyer, Alexander Kolesnikov, Dirk Weissenborn,
  Xiaohua Zhai, Thomas Unterthiner, Mostafa Dehghani, Matthias Minderer, Georg
  Heigold, Sylvain Gelly, Jakob Uszkoreit, and Neil Houlsby.
\newblock An image is worth 16x16 words: Transformers for image recognition at
  scale.
\newblock In \emph{International Conference on Learning Representations}, 2021.
\newblock URL \url{https://openreview.net/forum?id=YicbFdNTTy}.

\bibitem[Geirhos et~al.(2020)Geirhos, Jacobsen, Michaelis, Zemel, Brendel,
  Bethge, and Wichmann]{geirhos2020shortcut}
Robert Geirhos, J{\"o}rn-Henrik Jacobsen, Claudio Michaelis, Richard Zemel,
  Wieland Brendel, Matthias Bethge, and Felix~A Wichmann.
\newblock Shortcut learning in deep neural networks.
\newblock \emph{Nature Machine Intelligence}, 2\penalty0 (11):\penalty0
  665--673, 2020.

\bibitem[Hashimoto et~al.(2018)Hashimoto, Srivastava, Namkoong, and
  Liang]{hashimoto2018fairness}
Tatsunori Hashimoto, Megha Srivastava, Hongseok Namkoong, and Percy Liang.
\newblock Fairness without demographics in repeated loss minimization.
\newblock In \emph{International Conference on Machine Learning}, pages
  1929--1938. PMLR, 2018.

\bibitem[He et~al.(2016)He, Zhang, Ren, and Sun]{he2016deep}
Kaiming He, Xiangyu Zhang, Shaoqing Ren, and Jian Sun.
\newblock Deep residual learning for image recognition.
\newblock In \emph{Proceedings of the IEEE conference on computer vision and
  pattern recognition}, pages 770--778, 2016.

\bibitem[Hooker et~al.(2019)Hooker, Courville, Clark, Dauphin, and
  Frome]{hooker2019compressed}
Sara Hooker, Aaron Courville, Gregory Clark, Yann Dauphin, and Andrea Frome.
\newblock What do compressed deep neural networks forget?
\newblock \emph{arXiv preprint arXiv:1911.05248}, 2019.

\bibitem[Hui and Belkin(2021)]{hui2021evaluation}
Like Hui and Mikhail Belkin.
\newblock Evaluation of neural architectures trained with square loss vs
  cross-entropy in classification tasks.
\newblock In \emph{International Conference on Learning Representations}, 2021.
\newblock URL \url{https://openreview.net/forum?id=hsFN92eQEla}.

\bibitem[Idrissi et~al.(2022)Idrissi, Arjovsky, Pezeshki, and
  Lopez-Paz]{idrissi2022simple}
Badr~Youbi Idrissi, Martin Arjovsky, Mohammad Pezeshki, and David Lopez-Paz.
\newblock Simple data balancing achieves competitive worst-group-accuracy.
\newblock In \emph{Conference on Causal Learning and Reasoning}, pages
  336--351. PMLR, 2022.

\bibitem[Kirichenko et~al.(2023)Kirichenko, Izmailov, and
  Wilson]{kirichenko2022last}
Polina Kirichenko, Pavel Izmailov, and Andrew~Gordon Wilson.
\newblock Last layer re-training is sufficient for robustness to spurious
  correlations.
\newblock In \emph{The Eleventh International Conference on Learning
  Representations}, 2023.
\newblock URL \url{https://openreview.net/forum?id=Zb6c8A-Fghk}.

\bibitem[Liu et~al.(2015)Liu, Luo, Wang, and Tang]{liu2015deep}
Ziwei Liu, Ping Luo, Xiaogang Wang, and Xiaoou Tang.
\newblock Deep learning face attributes in the wild.
\newblock In \emph{Proceedings of the IEEE international conference on computer
  vision}, pages 3730--3738, 2015.

\bibitem[Maini et~al.(2023)Maini, Mozer, Sedghi, Lipton, Kolter, and
  Zhang]{maini2023can}
Pratyush Maini, Michael~C Mozer, Hanie Sedghi, Zachary~C Lipton, J~Zico Kolter,
  and Chiyuan Zhang.
\newblock Can neural network memorization be localized?
\newblock In \emph{Proceedings of the 40th International Conference on Machine
  Learning}, volume 202, pages 23536--23557. PMLR, 2023.

\bibitem[Qiu et~al.(2023)Qiu, Potapczynski, Izmailov, and
  Wilson]{qiu2023simple}
Shikai Qiu, Andres Potapczynski, Pavel Izmailov, and Andrew~Gordon Wilson.
\newblock Simple and fast group robustness by automatic feature reweighting.
\newblock In \emph{International Conference on Machine Learning}, pages
  28448--28467. PMLR, 2023.

\bibitem[Radford et~al.(2021)Radford, Kim, Hallacy, Ramesh, Goh, Agarwal,
  Sastry, Askell, Mishkin, Clark, et~al.]{radford2021learning}
Alec Radford, Jong~Wook Kim, Chris Hallacy, Aditya Ramesh, Gabriel Goh,
  Sandhini Agarwal, Girish Sastry, Amanda Askell, Pamela Mishkin, Jack Clark,
  et~al.
\newblock Learning transferable visual models from natural language
  supervision.
\newblock In \emph{International conference on machine learning}, pages
  8748--8763. PMLR, 2021.

\bibitem[Ribeiro et~al.(2016)Ribeiro, Singh, and Guestrin]{ribeiro2016should}
Marco~Tulio Ribeiro, Sameer Singh, and Carlos Guestrin.
\newblock ``why should i trust you?'' explaining the predictions of any
  classifier.
\newblock In \emph{Proceedings of the 22nd ACM SIGKDD international conference
  on knowledge discovery and data mining}, pages 1135--1144, 2016.

\bibitem[Sagawa et~al.(2020)Sagawa, Koh, Hashimoto, and
  Liang]{sagawa2019distributionally}
Shiori Sagawa, Pang~Wei Koh, Tatsunori~B. Hashimoto, and Percy Liang.
\newblock Distributionally robust neural networks.
\newblock In \emph{International Conference on Learning Representations}, 2020.
\newblock URL \url{https://openreview.net/forum?id=ryxGuJrFvS}.

\bibitem[Sohn(2016)]{sohn2016improved}
Kihyuk Sohn.
\newblock Improved deep metric learning with multi-class n-pair loss objective.
\newblock \emph{Advances in neural information processing systems}, 29, 2016.

\bibitem[Stephenson et~al.(2021)Stephenson, suchismita padhy, Ganesh, Hui,
  Tang, and Chung]{stephenson2021geometry}
Cory Stephenson, suchismita padhy, Abhinav Ganesh, Yue Hui, Hanlin Tang, and
  SueYeon Chung.
\newblock On the geometry of generalization and memorization in deep neural
  networks.
\newblock In \emph{International Conference on Learning Representations}, 2021.
\newblock URL \url{https://openreview.net/forum?id=V8jrrnwGbuc}.

\bibitem[Wah et~al.(2011)Wah, Branson, Welinder, Perona, and
  Belongie]{wah2011caltech}
C.~Wah, S.~Branson, P.~Welinder, P.~Perona, and S.~Belongie.
\newblock The caltech-ucsd birds-200-2011 dataset.
\newblock Technical Report CNS-TR-2011-001, California Institute of Technology,
  Computation \& Neural Systems Technical Report, 2011.

\bibitem[Yang et~al.(2023)Yang, Nushi, Palangi, and
  Mirzasoleiman]{yang2023mitigating}
Yu~Yang, Besmira Nushi, Hamid Palangi, and Baharan Mirzasoleiman.
\newblock Mitigating spurious correlations in multi-modal models during
  fine-tuning.
\newblock In \emph{International Conference on Machine Learning}, pages
  39365--39379. PMLR, 2023.

\end{thebibliography}

\clearpage
\appendix

\clearpage
\newpage
\clearpage
\begin{center}
{\bf {\Large Supplementary materials for\\[3mm]

{\bf {Uncovering Memorization Effect in the Presence of Spurious Correlations}}\\[3mm]}}
\end{center}
\appendix

\begin{table}[ht]
\caption{\textbf{Data Augmentation}. This table lists the detailed setting for the data augmentations adopted in the image preprocessing for the experiments conducted in Section~\ref{sec:stage 2}.} 
\label{table:augmentation}
\centering
{
\begin{tabular}{@{\hskip 0.1mm}lccccc@{\hskip 0.1mm}}
\toprule
\textbf{Class}  && \textbf{Parameter} && \textbf{Value} \\ 
\midrule
\multirow{3}{*}{RandomResizedCrop}  &&  target\_size  &&  (224,224) \\
  &&  scale &&  (0.7,1.0) \\
  &&  ratio &&   (0.75,1.33) \\
\midrule
\multirow{1}{*}{RandomHorizontalFlip}  &&  $p$  &&  0.5 \\
\midrule
\multirow{1}{*}{RandomRotation}  &&  degrees  &&  15 \\
\midrule
\multirow{3}{*}{RandomAffine}  &&  degrees  &&  0 \\
  &&  translate &&  (0.1,0.1) \\
  &&  ratio &&   (0.9,1.1) \\
\midrule
\multirow{1}{*}{RandomPerspective}  &&  distortion\_scale  &&  0.2 \\
\bottomrule
\end{tabular}
}
\end{table}

\begin{figure}[ht]
\begin{center}
\includegraphics[width=0.48\columnwidth]{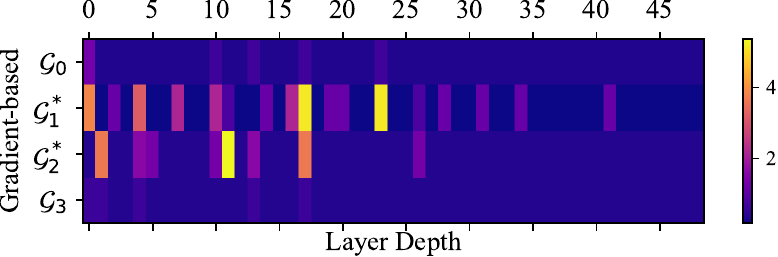}
\includegraphics[width=0.48\columnwidth]{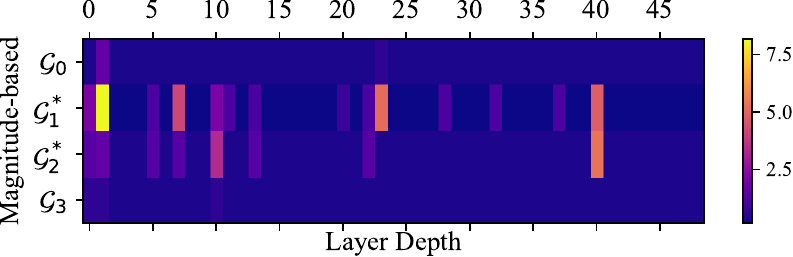}
\caption{\textbf{Group accuracy change by \textit{only} prunning top-3 largest gradient or magnitude neurons within each conv layer in {ResNet-50}}. The color bar is in the scale of percentage.}
\label{fig:structured heatmap}
\end{center}
% \vspace{-10pt}
\end{figure}

\begin{figure}[ht]
\begin{center}
\includegraphics[width=0.48\columnwidth]{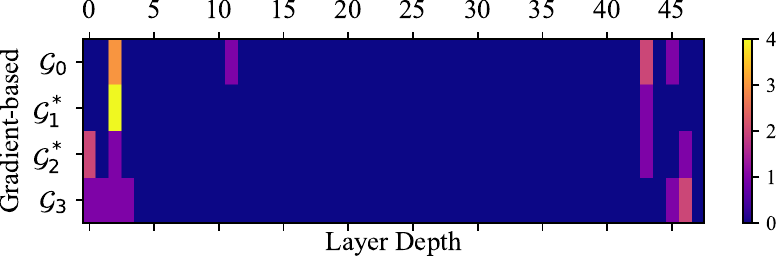}
\caption{\textbf{Distribution of Critical Neurons in ResNet-50 via Unstructured Tracing.} This analysis utilizes a gradient-based criterion to identify critical neurons, with the color bar indicating the neuron count.} 
\label{fig:unstructured heatmap}
\end{center}
\end{figure}

\begin{figure*}[ht]
\begin{center}
\centerline{
\includegraphics[width=\linewidth]{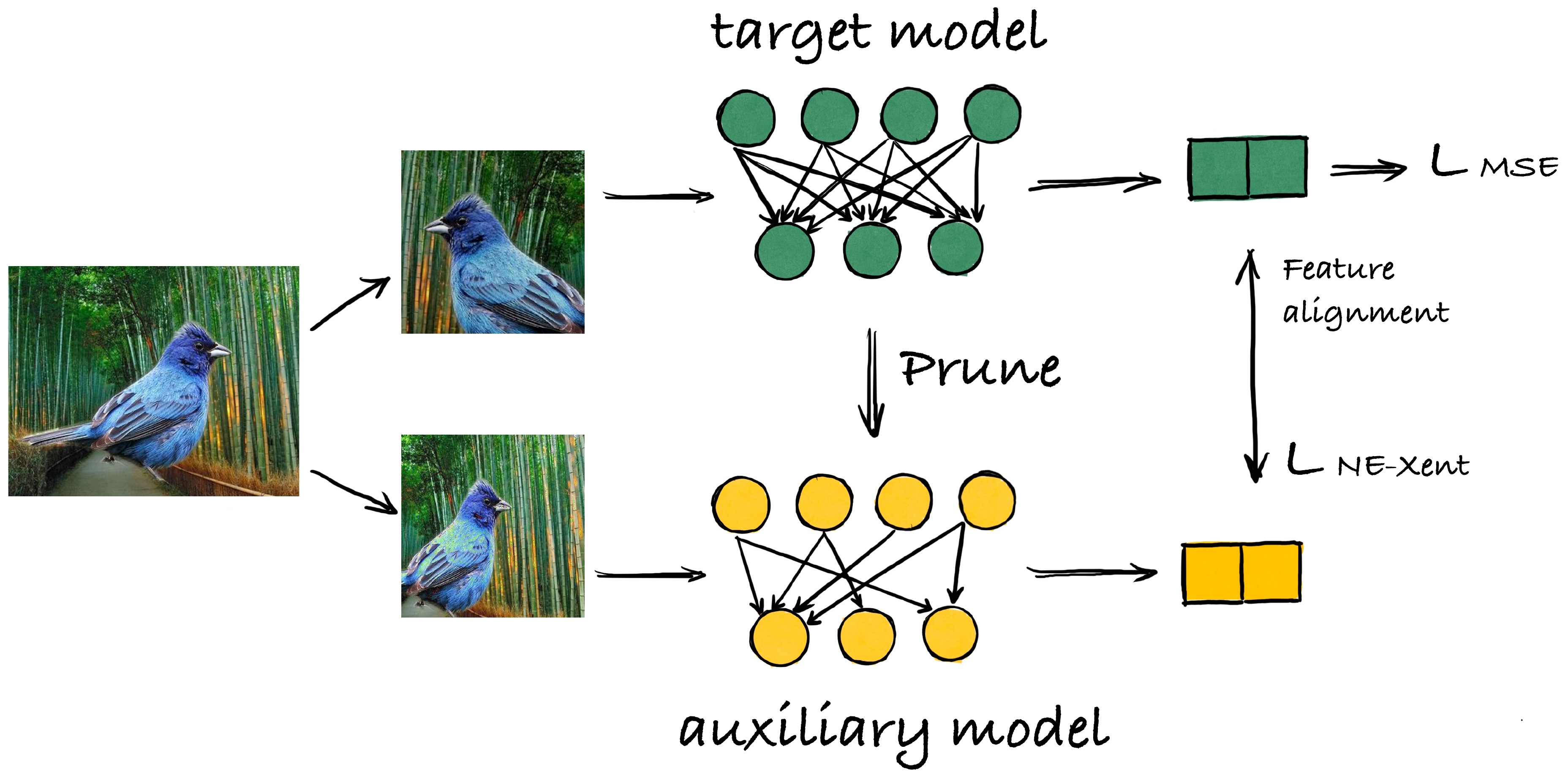}
}
\caption{\textbf{The overview of our proposed fine-tuning framework.}
The key innovation of our framework lies in its dual-branch architecture, consisting of the target model and an auxiliary model, the latter being a pruned version of the target with critical neurons removed 
The weights of non-pruned neurons are shared between both models. 
At the beginning of each fine-tuning epoch, the auxiliary model inherits weights from the latest target model, and both models are updated concurrently during training. 
By contrasting the two models using the NT-Xent loss~\citep{sohn2016improved}, we implicitly counteract the spurious memorization attributed to the critical neurons.}
\label{fig:model}
\end{center}
\end{figure*}

\begin{table}[ht]
\caption{\textbf{Experimental Settings}.} 
\label{table:hyperparameter}
% \vspace{-5pt}
\centering
\resizebox{0.83\textwidth}{!}{
% \begin{tabular}{@{\hskip 1mm}l|ccccc@{\hskip 1mm}}
\begin{tabular}{@{\hskip 0.1mm}lccccc@{\hskip 0.1mm}}
\toprule
\textbf{Condition}  && \textbf{Parameter} && \textbf{Value} \\ 
\midrule
% \midrule
\textit{Model Architecture}: && && \\
\midrule
\multirow{3}{*}{ResNet-50~\cite{he2016deep}} &&  Input size  &&  {224$\times$224} \\
  &&  Output size of projection layer &&   2 \\
\midrule
\multirow{3}{*}{ViT-S/16@224px~\cite{dosovitskiy2020image}}  && Input size &&  {224$\times$224} \\
  && Output size of projection layer &&  2 \\
\midrule
% \midrule
\textit{ERM Training}: && && \\
\midrule
\multirow{5}{*}{Optimizer}  &&  Type  &&  Adam \\
  &&  Learning rate &&  1e-4 \\
  &&  Momentum &&   0.9 \\
  &&  L2 weight decay &&   0 \\
  && Metric to pick best model && WGA\\
\midrule
\multirow{3}{*}{Scheduler}  &&  Type  &&  ReduceLROnPlateau \\
  &&  Factor &&  0.5 \\
  &&  Patience &&   3 \\
\midrule
\multirow{1}{*}{Criterion}  &&  Type &&  Cross Entropy Loss \\
\midrule
\multirow{1}{*}{Batch size}  &&   &&  256 \\
\midrule
\midrule
\textit{Finetuning}: && && \\
\midrule
\multirow{5}{*}{Optimizer}  &&  Type  &&  Adam \\
  &&  Learning rate &&  2e-4 \\
  &&  Momentum &&   0.9 \\
  &&  L2 weight decay &&   0 \\
  && Metric to pick best model && WGA\\
\midrule
\multirow{3}{*}{Scheduler}  &&  Type  &&  ReduceLROnPlateau \\
  &&  Factor &&  0.5 \\
  &&  Patience &&   1 \\
\midrule
\multirow{3}{*}{  Ours (gradient-based)}  &&  Pruning percentage  &&  0.01\% \\
  &&  Loss function &&  NTXent$+$MSE \\
  &&  $\lambda$ &&   0.2 \\
\midrule
\multirow{3}{*}{  Ours (magnitude-based)}  &&  Pruning percentage  &&  0.01\% \\
  &&  Loss function &&  NTXent$+$MSE \\
  &&  $\lambda$ &&   0.2 \\
\midrule
\textit{Dataset-specific}: && && \\
\midrule
Waterbirds~\cite{wah2011caltech, sagawa2019distributionally}  &&  Input size  &&  $224\times224$ \\
CelebA~\cite{liu2015deep}  &&  Input size  &&  $224\times224$ \\
\bottomrule
\end{tabular}
}
\end{table}

\begin{table}[ht]
\caption{\textbf{Ablation on Number of Pruned Neurons.} 
We compare different choice of numbers of pruned neurons, and then calculate the drop in the training accuracy for each group in Waterbirds. 
The selection adopts the magnitude-based criterion. 
}
\label{table:why top 3}

\centering
\resizebox{0.34\textwidth}{!}{
\begin{tabular}{@{\hskip 1mm}lcccc@{\hskip 1mm}}
\toprule
& \multicolumn{4}{c}{\textbf{Change in group training accuracy}}\\
\cmidrule(r){2-5}
top-$k$  & $\mathcal{G}_0$ & $\mathcal{G}_1$ & $\mathcal{G}_2$ & $\mathcal{G}_3$  \\ 
\midrule
1 &      1.00\% &    4.90\% &   3.57\% &   0.38\% \\
\midrule
2 &   0.57\% &   3.07\% &   7.15\% &   0.18\% \\
\midrule
3 &   0.94\% &   1.63\% &  10.72\% &   0.28\% \\
\midrule
5 &    3.70\% &  20.72\% &   4.55\% &   1.26\% \\
\midrule
10 &   4.06\% &  20.01\% &   9.07\% &   1.42\% \\
\bottomrule
\end{tabular}
}
\end{table}

\begin{table}[ht]
\caption{\textbf{Ablation on the Loss Functions.} 
We compare MSE with Cross Entropy (CE) as the choice of the loss term in the model fine-tuning. 
}
\label{table:loss function}

\centering
\resizebox{0.73\textwidth}{!}{
\begin{tabular}{@{\hskip 1mm}lcccc@{\hskip 1mm}}
\toprule
\multicolumn{3}{c}{\textbf{Experimental Setup}} & \multicolumn{2}{c}{\textbf{Group Accuracy}}\\
\cmidrule(r){1-3} \cmidrule(r){4-5}
Training Strategy  & Magnitude Pruning Percentage & Loss Term & WGA & Difference from ERM  \\ 
\midrule
ERM & N/A & CE & 64.02\% & N/A \\
\midrule
\multirow{4}{*}{ERM+Fine-tuning (ours)} & 0.01\%	& CE &71.34\%	&7.32\%\\
&0.01\%	& MSE &	80.89\% &16.87\% \\
&0.10\%	& CE & 77.26\% & 13.24\% \\
&0.10\%	& MSE & 79.91\% & 15.89\%
\\
% \midrule
% JTT~\cite{liu2021just} & 61.68 & 90.63 & 83.64 & 97.29
% \\
\bottomrule
\end{tabular}
}
\end{table}

\begin{table}[ht]
\caption{\textbf{Ablation on Kick-in Epochs.} 
We compare different choice of the kick-in epoch. The kick-in epoch refers to the number of epochs for ERM pretraining the target model before the fine-tuning stage.
In this experiment, other experimental setup is as follows: percentage of magnitude pruning $=$ 0.01\% (\ie~pruning the top 0.01\% largest neurons); MSE loss term; loss term ratio $\lambda=0.2$ in \eqref{eq:total loss}; fine-tuning for 20 epochs after kick-in.
}
\label{table:kickin epoch}

\centering
\resizebox{0.54\textwidth}{!}{
\begin{tabular}{@{\hskip 1mm}lccc@{\hskip 1mm}}
\toprule
\multicolumn{2}{c}{\textbf{Experimental Setup}} & \multicolumn{2}{c}{\textbf{Group Accuracy}}\\
\cmidrule(r){1-2} \cmidrule(r){3-4}
Training Strategy  & Kick-in Epoch & WGA & Difference from ERM  \\ 
\midrule
ERM & N/A & 64.02\% &  N/A \\
\midrule
\multirow{4}{*}{ERM+Fine-tuning (ours)} & 30 &65.89\% &1.87\%\\
& 40 & 80.89\% & 16.87\%  \\
& 50 & 73.68\% & 9.66\%  \\
& 60 & 76.23\% & 12.21\% 
\\
% \midrule
% JTT~\cite{liu2021just} & 61.68 & 90.63 & 83.64 & 97.29
% \\
\bottomrule
\end{tabular}
}
\end{table}

\begin{table}[ht]
\caption{\textbf{Ablation on Number of Fine-tuning Epochs.} 
We compare different numbers of epochs for fine-tuning. The kick-in epoch is fixed at 40.
Other experimental setup is as follows: percentage of magnitude pruning  at 0.01\%; MSE loss term; loss term ratio $\lambda=0.2$ in Eq.~\eqref{eq:total loss}.
}
\label{table:finetuning epochs}

\centering
\resizebox{0.57\textwidth}{!}{
\begin{tabular}{@{\hskip 1mm}lccc@{\hskip 1mm}}
\toprule
\multicolumn{2}{c}{\textbf{Experimental Setup}} & \multicolumn{2}{c}{\textbf{Group Accuracy}}\\
\cmidrule(r){1-2} \cmidrule(r){3-4}
Training Strategy  & Fine-tuning Epochs & WGA & Difference from ERM  \\ 
\midrule
ERM & N/A & 64.02\% &  N/A \\
\midrule
\multirow{3}{*}{ERM+Fine-tuning (ours)} & 10 & 54.98\% & -9.04\%\\
& 20 & 80.89\% & 16.87\%  \\
& 30 & 80.89\% & 16.87\%  \\
% & 40 & 80.89 & 16.87\% \\
% \midrule
% JTT~\cite{liu2021just} & 61.68 & 90.63 & 83.64 & 97.29
% \\
\bottomrule
\end{tabular}
}
\end{table}

\begin{table}[ht]
\caption{\textbf{Ablation on Gradient Sources.} 
We compare two source of calculating the gradient in the gradient-based criterion for critical neuron selection: the worst predicted examples from the full training set versus from the subset of minority examples. 
}
\label{table:gradient source}

\centering
\resizebox{0.77\textwidth}{!}{
\begin{tabular}{@{\hskip 1mm}lcccc@{\hskip 1mm}}
\toprule
\multicolumn{3}{c}{\textbf{Experimental Setup}} & \multicolumn{2}{c}{\textbf{Group Accuracy}}\\
\cmidrule(r){1-3} \cmidrule(r){4-5}
Training Strategy  & Gradient Pruning Percentage & Gradient Source & WGA & Difference from ERM  \\ 
\midrule
ERM & N/A & CE & 64.02\% & N/A \\
\midrule
\multirow{4}{*}{ERM+Fine-tuning (ours)} & 0.01\%	& Minority Groups & 74.45\%	& 10.43\%\\
&0.01\%	& Full Training Set &	76.64\% & 12.62\% \\
&0.02\%	& Minority Groups& 65.89\% & 1.89\% \\
&0.02\%	& Full Training Set & 78.50\% & 14.48\%
\\

\bottomrule
\end{tabular}
}
\end{table}

\begin{table}[ht]
\caption{\textbf{Ablation on Loss term Ratios.} 
We compare different choice of the loss term ratio $\lambda$ in Eq.~\eqref{eq:total loss}.
Other experimental setup is as follows: percentage of magnitude pruning $=$ 0.01\% (\ie~pruning the top 0.01\% largest neurons); MSE loss term; the number of fine-tuning epochs $=20$.
}
\label{table:loss ratio}

\centering
\resizebox{0.47\textwidth}{!}{
\begin{tabular}{@{\hskip 1mm}lccc@{\hskip 1mm}}
\toprule
\multicolumn{2}{c}{\textbf{Experimental Setup}} & \multicolumn{2}{c}{\textbf{Group Accuracy}}\\
\cmidrule(r){1-2} \cmidrule(r){3-4}
Training Strategy  & $\lambda$ & WGA & Difference from ERM  \\ 
\midrule
ERM & N/A & 64.02\% &  N/A \\
\midrule
\multirow{4}{*}{ERM+Fine-tuning (ours)} & 0.01 & 65.26\% &1.24\%\\
& 0.2 & 80.89\% & 16.87\%  \\
& 1 & 72.43\% & 8.41\%  \\
& 10 & 59.97\% & -4.05\% 
\\
% \midrule
% JTT~\cite{liu2021just} & 61.68 & 90.63 & 83.64 & 97.29
% \\
\bottomrule
\end{tabular}
}
\end{table}

\begin{table}[ht]
\caption{\textbf{Ablation on Pruning Percentage.} 
We compare different choice of pruning percentage.
Other experimental setup is as follows: percentage of magnitude pruning $=$ 0.01\% (\ie, pruning the top 0.01\% largest neurons); MSE loss term; the number of fine-tuning epochs $=20$.
}
\label{table:pruning percentage}

\centering
\resizebox{0.70\textwidth}{!}{
\begin{tabular}{@{\hskip 1mm}lcccc@{\hskip 1mm}}
\toprule
\multicolumn{2}{c}{\textbf{Experimental Setup}} & \multicolumn{2}{c}{\textbf{Group Accuracy}}\\
\cmidrule(r){1-2} \cmidrule(r){3-4}
Training Strategy  & Criterion & Pruning Percentage & WGA & Difference from ERM  \\ 
\midrule
ERM & N/A & N/A & 64.02\% &  N/A \\
\midrule
\multirow{4}{*}{ERM+Fine-tuning (ours)} & \multirow{4}{*}{Gradient-based} & 0.01\% & 76.64\% &12.62\%\\
& & 0.1\% & 77.26\% & 13.24\%  \\
& & 1\% & 71.50\% & 7.48\%  \\
& & 10\% & 0.04\% & -63.98\%  \\
% & & 20\% & 72.43\% & 8.41\%  \\
% & & 30\% & 59.97\% & -4.05\% 
\midrule
\multirow{4}{*}{ERM+Fine-tuning (ours)} & \multirow{4}{*}{Magnitude-based} & 0.01\% & 80.89\% &16.87\%\\
& & 0.1\% & 79.91 & 15.89\%  \\
& & 1\% & 69.00\% & 4.98\%  \\
& & 10\% & 73.83\% & 9.81\%  \\
% & & 20\% & 72.43\% & 8.41\%  \\
% & & 30\% & 59.97\% & -4.05\% 
% JTT~\cite{liu2021just} & 61.68 & 90.63 & 83.64 & 97.29
% \\
\bottomrule
\end{tabular}
}
\end{table}

\begin{table}[ht]
\caption{\textbf{Ablation on the Mixed Pruning Criteria.} 
We test a mixed pruning criterion that combines the gradient-based and magnitude-based ones. 
Other experimental setup is as follows: percentage of magnitude pruning $=$ 0.01\% (\ie~pruning the top 0.01\% largest neurons); MSE loss term; fine-tuning epochs $=$ 20; gradient source is the full training set. 
}
\label{table:combining criteria}

\centering
\resizebox{0.86\textwidth}{!}{
\begin{tabular}{@{\hskip 1mm}lcccc@{\hskip 1mm}}
\toprule
\multicolumn{3}{c}{\textbf{Experimental Setup}} & \multicolumn{2}{c}{\textbf{Group Accuracy}}\\
\cmidrule(r){1-3} \cmidrule(r){4-5}
Training Strategy  & Magnitude Pruning Percentage & Gradient Pruning Percentage & WGA & Difference from ERM  \\ 
\midrule
ERM & N/A & N/A & 64.02\% & N/A \\
\midrule
\multirow{4}{*}{ERM+Fine-tuning (ours)} & 0.01\% & 0.01\%  &77.26\%	& 13.24\%\\
&0.1\%	&0.1\%& 	77.57\% & 13.55\% \\
&1\%	&1\%& 69.22\% & 5.20\% \\
&10\%	&10\%& 0.18\% & -63.84\%
\\

\bottomrule
\end{tabular}
}
\end{table}

\begin{table}[ht]
\caption{\textbf{Unstructured Tracing on CelebA.} 
We conduct the additional unstructured tracing experiment on CelebA. The table presents the changes in training accuracy for each group. 
Note that for the CelebA dataset, the minority group is denoted as $\mathcal{G}_3^*$, with the star superscript indicating its minority status.
}
% \caption{\red{\textbf{Unstructured Tracing on CelebA.} 
% We conduct the additional unstructured tracing experiment on CelebA. The table presents the changes in training accuracy for each group. 
% Note that for the CelebA dataset, the minority group is denoted as $\mathcal{G}_3^*$, with the star superscript indicating its minority status.
% }}
\label{table:celebA unstructured train}

\centering
\resizebox{0.5\textwidth}{!}{
\begin{tabular}{@{\hskip 1mm}lccccccc@{\hskip 1mm}}
\toprule
\multicolumn{3}{c}{\textbf{Experimental Setup}} & \multicolumn{4}{c}{\textbf{Group Accuracy \% Change}}\\
\cmidrule(r){1-3} \cmidrule(r){4-7}
Modification  & top-$k$ & Std. Dev. & $\mathcal{G}_0$ & $\mathcal{G}_1$ & $\mathcal{G}_2$ & $\mathcal{G}_3^*$  
\\
\midrule
zero-out & 1&n/a&2.64&0.26&4.85&10.24
\\
zero-out &2&n/a&2.63&0.23&4.77&9.23
\\
zero-out &3&n/a&3.34&0.26&5.29&9.66
\\ 
\midrule
random init	& 1& 0.005 & 2.68& 0.26& 4.86 & 10.34
\\
random init	& 2	& 0.005	& 2.59& 0.23 &	4.75 & 9.16 
\\
random init	& 3 & 0.005	&3.39&0.28&5.32&9.93
\\
random init&1&0.01&2.81&0.28&4.94&10.86
\\
random init&2&0.01&2.87&0.26&4.96&9.98
\\
random init&3&0.01&3.30&0.25&5.17&9.21
\\
random init&1&0.02&2.88&0.30&4.97&11.13
\\
random init&2&0.02&2.54&0.23&4.51&8.53
\\
random init&3&0.02&3.57&0.31&5.37&10.82
\\

\bottomrule
\end{tabular}
}
\end{table}

\begin{table}[ht]
% \caption{\red{\textbf{Summary of Experiments in Sec.\ref{sec:stage2 result}}. 
% The mean and standard deviation of Worst Group Accuracy (WGA) of our proposed method in Sec.~\ref{sec:stage2 result} are included in this table. 
% The statistics is computed over 10 random seeds.
% The mean values here are the same as the reported numbers in Figure~\ref{fig:wga comparison}. 
% }}
\caption{\textbf{Summary of Experiments in Sec. \ref{sec:stage2 result}.} This table presents the mean and standard deviation of Worst Group Accuracy (WGA) for our proposed method, computed over 10 random seeds. The mean values reported here are consistent with those shown in Figure \ref{fig:wga comparison}.}
% \caption{\red{\textbf{Summary of Experiments in Sec. \ref{sec:stage2 result}.} This table presents the mean and standard deviation of Worst Group Accuracy (WGA) for our proposed method, computed over 10 random seeds. The mean values reported here are consistent with those shown in Figure \ref{fig:wga comparison}.}}
\label{table:summary statistics}

\centering
\resizebox{0.5\textwidth}{!}{
\begin{tabular}{@{\hskip 1mm}lccc@{\hskip 1mm}}
\toprule
& \multicolumn{3}{c}{\textbf{WGA (\%)}}\\
\cmidrule(r){2-4}
\textbf{Setting} & \textbf{ERM} & \textbf{Gradient} & \textbf{Magnitude} \\
\midrule
ResNet on Waterbirds & 64.0$\pm$0.45 & 77.3$\pm$4.91 & 80.9$\pm$4.36 \\
ResNet on CelebA & 47.0$\pm$1.25 & 57.8$\pm$3.58 & 55.6$\pm$2.01 \\
ViT on Waterbirds & 52.7$\pm$2.79 & 69.6$\pm$5.52 & 76.5$\pm$5.74 \\
ViT on CelebA & 47.2$\pm$1.59 & 56.7$\pm$3.74 & 52.8$\pm$1.32 \\
% \midrule
% JTT~\cite{liu2021just} & 61.68 & 90.63 & 83.64 & 97.29
% \\
\bottomrule
\end{tabular}
}
\end{table}

\begin{table}[ht]
\caption{\textbf{Comparison between Changes in Training Accuracy and Changes in Test Accuracy.} 
We compare the percentage change in the training accuracy and test accuracy on Waterbirds. 
The results show that for the minority groups $\mathcal{G}_1^*$ and $\mathcal{G}_2^*$ under almost all experimental setups, the test accuracy is affected significantly less than the training accuracy when modifying the critical neurons. 
Specifically, the test accuracy for minority groups remains quite stable after modifying critical neurons, as opposed to the training accuracy. This result provides additional evidence for the memorization role of the critical neurons.
Note that for the Waterbirds dataset, the minority group is denoted as $\mathcal{G}_1^*$ and $\mathcal{G}_2^*$, with the star superscript indicating their minority status.
}
% \caption{\red{\textbf{Comparison between Changes in Training Accuracy and Changes in Test Accuracy.} 
% We compare the percentage change in the training accuracy and test accuracy on Waterbirds. 
% The results show that for the minority groups $\mathcal{G}_1^*$ and $\mathcal{G}_2^*$ under almost all experimental setups, the test accuracy is affected significantly less than the training accuracy when modifying the critical neurons. 
% Specifically, the test accuracy for minority groups remains quite stable after modifying critical neurons, as opposed to the training accuracy. This result provides additional evidence for the memorization role of the critical neurons.
% Note that for the Waterbirds dataset, the minority group is denoted as $\mathcal{G}_1^*$ and $\mathcal{G}_2^*$, with the star superscript indicating their minority status.
% }}
\label{table:compare train test waterbirds}

\centering
\resizebox{0.85\textwidth}{!}{
\begin{tabular}{@{\hskip 1mm}lcccccccccc@{\hskip 1mm}}
\toprule
\multicolumn{3}{c}{\textbf{Experimental Setup}} & \multicolumn{4}{c}{\textbf{Group Accuracy \% Change (training set)}} &
\multicolumn{4}{c}{\textbf{Group Accuracy \% Change (test set)}}
\\
\cmidrule(r){1-3} \cmidrule(r){4-7} \cmidrule(r){8-11}

Modification  & Top-$k$ & Std. Dev. & $\mathcal{G}_0$ & $\mathcal{G}_1^*$ & $\mathcal{G}_2^*$ & $\mathcal{G}_3$  & $\mathcal{G}_0$ & $\mathcal{G}_1^*$ & $\mathcal{G}_2^*$ & $\mathcal{G}_3$  
\\
\midrule
zero-out & 1 & N/A & 1.00 & 4.90 & 3.57 & 0.38 & 0.93 & {0.89} & 1.25 & 0.16 \\
zero-out & 2 & N/A & 0.57 & 3.07 & 7.15 & 0.18 & 1.51 & 2.48 & 0.31 & 0.62 \\
zero-out & 3 & N/A & 0.94 & 1.63 & 10.72 & 0.28 & 1.77 & {2.17} & 0.31 & 0.16 \\
\midrule 
random noise & 1 & 0.005 & 0.08 & 0.34 & 0.67 & 0.01 & 0.07 & 0.31 & 0.12 & 0.03 \\
random noise & 2 & 0.005 & 0.08 & 0.78 & 0.89 & 0.03 & 0.09 & 0.23 & 0.22 & 0.05 \\
random noise & 3 & 0.005 & 0.06 & 0.76 & 0.89 & 0.03 & 0.08 & 0.35 & 0.22 & 0.07 \\
random noise & 1 & 0.01 & 0.06 & 0.61 & 0.89 & 0.02 & 0.07 & 0.39 & 0.29 & 0.02 \\
random noise & 2 & 0.01 & 0.07 & 0.60 & 0.20 & 0.02 & 0.08 & 0.40 & {0.34} & 0.12 \\
random noise & 3 & 0.01 & 0.03 & 1.52 & 0.71 & 0.03 & 0.16 & 0.27 & 0.22 & 0.24 \\
random noise & 1 & 0.02 & 0.09 & 1.29 & 0.67 & 0.06 & 0.13 & 0.54 & 0.16 & 0.16 \\
random noise & 2 & 0.02 & 0.11 & 1.44 & 1.59 & 0.01 & 0.17 & 0.49 & 0.22 & 0.28 \\
random noise & 3 & 0.02 & 0.09 & 1.84 & 2.32 & 0.06 & 0.31 & 0.47 & 0.29 & 0.33 \\

\bottomrule
\end{tabular}
}
\end{table}

\begin{table}[ht]
\caption{\textbf{Raw Values of Change in the Group Accuracy}. 
This table provides the raw values for the changes in group accuracy corresponding to the experimental results in Figures~\ref{fig:random initialization barplot} and \ref{fig:random noise barplot}. It clearly demonstrates that for minority group $G_1^*$ and $G_2^*$ training accuracy consistently decreases across all experimental settings and for every choice of $k$ in top-$k$.
Note that for the Waterbirds dataset, the minority group is denoted as $\mathcal{G}_1^*$ and $\mathcal{G}_2^*$.
}
% \caption{\red{\textbf{Raw Values of Change in the Group Accuracy}. 
% This table provides the raw values for the changes in group accuracy corresponding to the experimental results in Figures~\ref{fig:random initialization barplot} and \ref{fig:random noise barplot}. It clearly demonstrates that for minority group $G_1^*$ and $G_2^*$ training accuracy consistently decreases across all experimental settings and for every choice of $k$ in top-$k$.
% Note that for the Waterbirds dataset, the minority group is denoted as $\mathcal{G}_1^*$ and $\mathcal{G}_2^*$.
% }}
\label{table:raw change}
\centering
\resizebox{0.5\textwidth}{!}{
\begin{tabular}{@{\hskip 1mm}lccccc@{\hskip 1mm}}
\toprule
& & \multicolumn{4}{c}{\textbf{Group Accuracy \% Change}}\\
\cmidrule(r){3-6}
& top-$k$ & $\mathcal{G}_0$ & $\mathcal{G}_1^*$ & $\mathcal{G}_2^*$ & $\mathcal{G}_3$ \\
\midrule
% \multicolumn{5}{l}{Figure~\ref{fig:random initialization barplot} (left)} \\
% \midrule
        \multirow{3}{*}{Figure~\ref{fig:random initialization barplot} (left)}&top1 & 1.25 & -8.92 & -6.08 & -0.18 \\
        &top2 & 1.16 & -11.80 & -6.97 & 1.83 \\
        &top3 & -0.36 & -6.80 & -3.04 & 1.71 \\
\midrule
% \multicolumn{5}{l}{Figure~\ref{fig:random initialization barplot} (right)} \\
% \midrule
        \multirow{3}{*}{Figure~\ref{fig:random initialization barplot} (right)}&top1 & -1.30 & -6.95 & -2.19 & -0.18 \\
        &top2 & 0.90 & -2.84 & -3.38 & 0.10 \\
        &top3 & -2.63 & -10.03 & -3.54 & -0.37 \\
\midrule
% \multicolumn{5}{l}{Figure~\ref{fig:random noise barplot} (left)} \\
% \midrule
        \multirow{3}{*}{Figure~\ref{fig:random noise barplot} (left)}&top1 & -0.07 & -0.30 & -0.6 & -0.01 \\
        &top2 & -0.08 & -0.78 & -0.89 & -0.03 \\
        &top3 & -0.07 & -0.84 & -0.89 & -0.04 \\
\midrule
% \multicolumn{5}{l}{Figure~\ref{fig:random noise barplot} (right)} \\
% \midrule
        \multirow{3}{*}{Figure~\ref{fig:random noise barplot} (right)}&top1 & -0.86 & -3.20 & -3.97 & 0.30 \\
        &top2 & -0.59 & -0.18 & -6.75 & -0.17 \\
        &top3 & -0.89 & -1.09 & -9.13 & -0.28 \\
\bottomrule
\end{tabular}
}
\end{table}

\begin{figure}[ht]
    \centering
    \includegraphics[height=0.12\textheight]{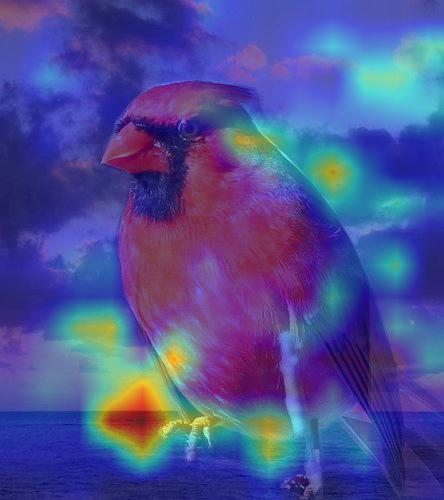}
    \includegraphics[height=0.12\textheight]{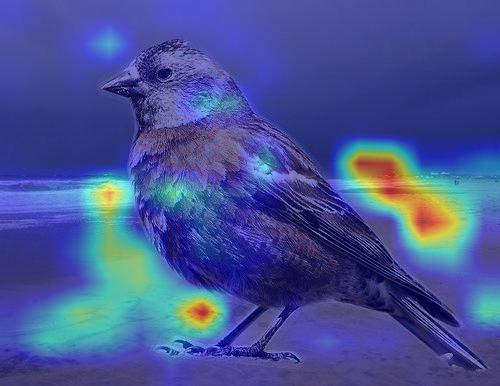}
    \includegraphics[height=0.12\textheight]{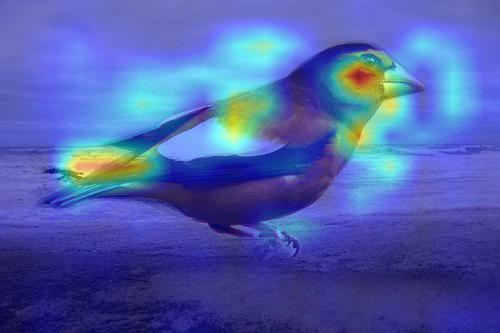}
    \includegraphics[height=0.12\textheight]{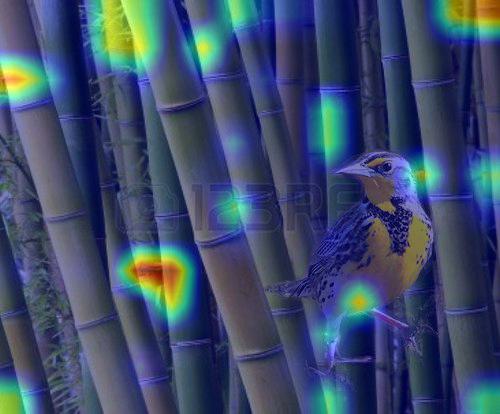}
    \includegraphics[height=0.12\textheight]{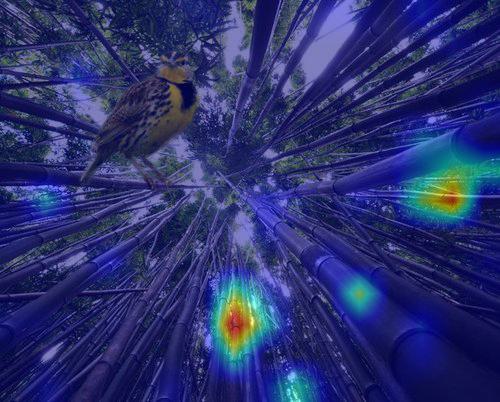}
    \includegraphics[height=0.12\textheight]{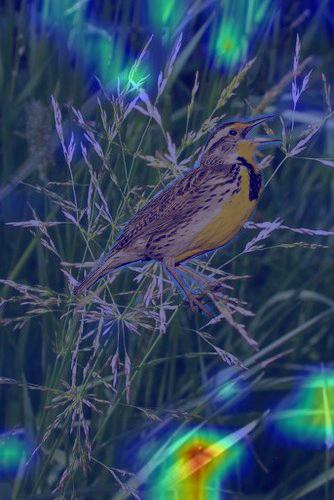}
    \includegraphics[height=0.12\textheight]{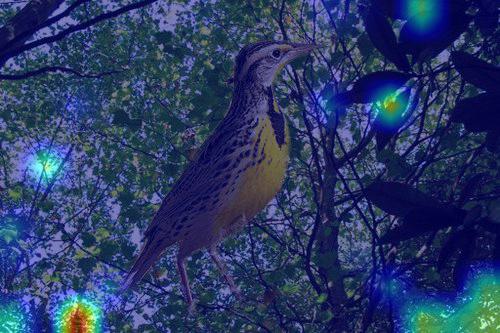}
    \includegraphics[height=0.12\textheight]{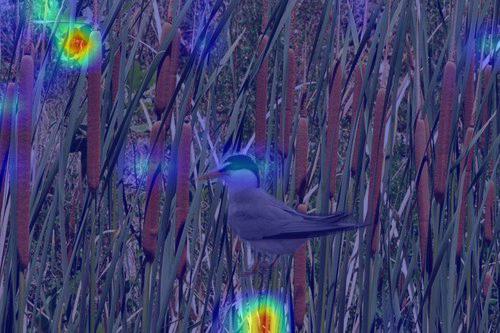}
    
    \includegraphics[height=0.12\textheight]{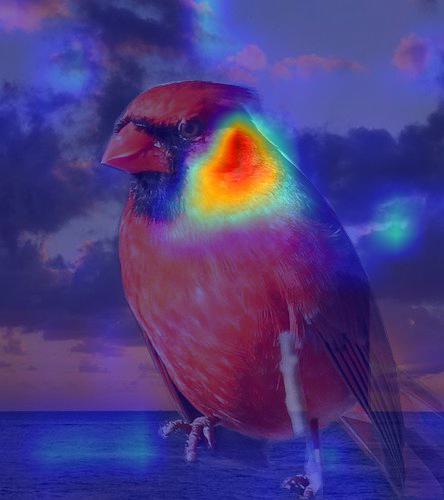}
    \includegraphics[height=0.12\textheight]{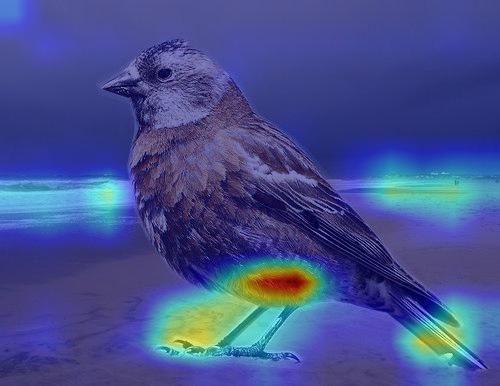}
    \includegraphics[height=0.12\textheight]{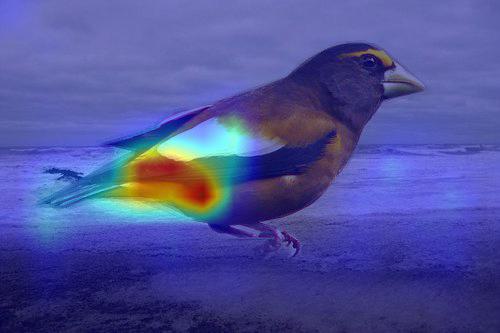}
    \includegraphics[height=0.12\textheight]{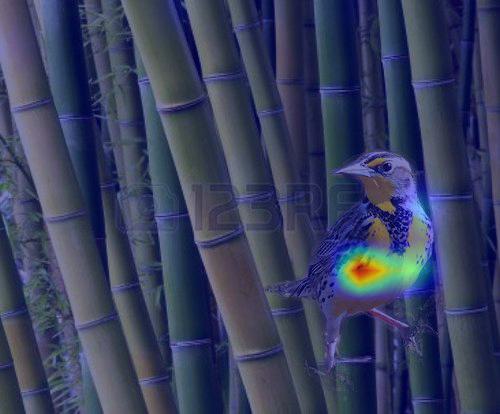}
    \includegraphics[height=0.12\textheight]{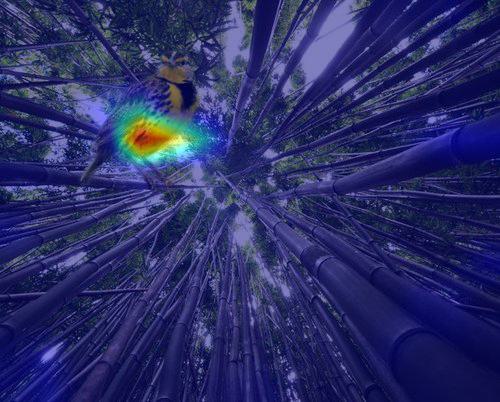}
    \includegraphics[height=0.12\textheight]{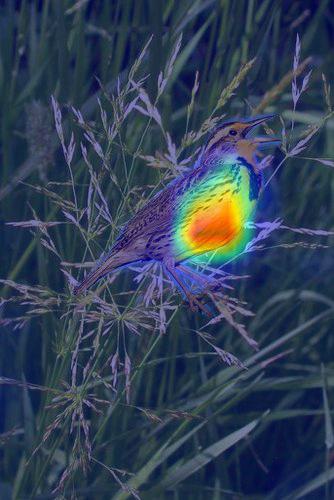}
    \includegraphics[height=0.12\textheight]{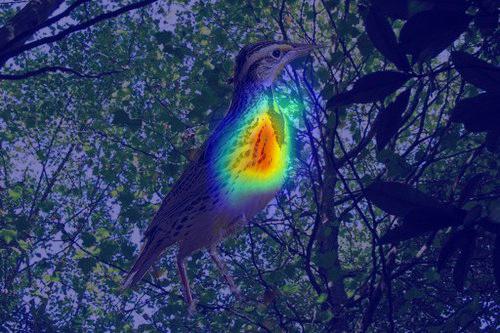}
    \includegraphics[height=0.12\textheight]{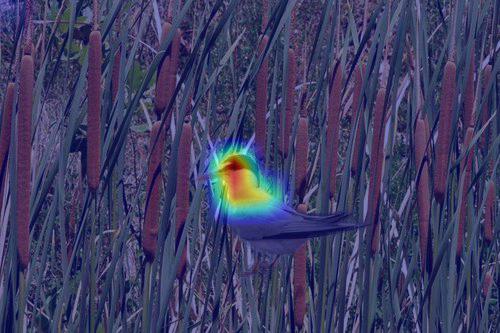}
    \caption{{GradCAM visualization} of Waterbirds on ResNet-50 by ERM (top two rows) and our fine-tuning strategy (bottom two rows).}
    \label{fig:gradcam}
\end{figure}

\begin{figure}
    \centering
    \includegraphics[width=0.25\textwidth]{figs/random_init_largest_activation_0.005_barplot.pdf}
    \includegraphics[width=0.25\textwidth]{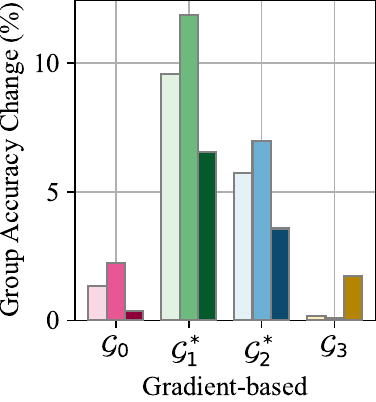}
    \includegraphics[width=0.25\textwidth]{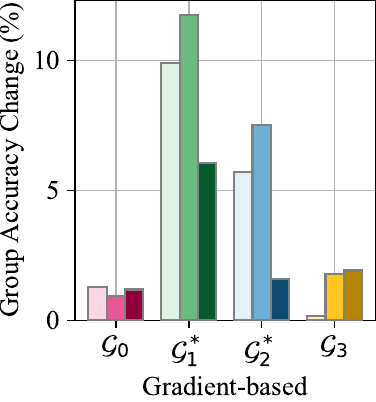}
    \includegraphics[width=0.25\textwidth]{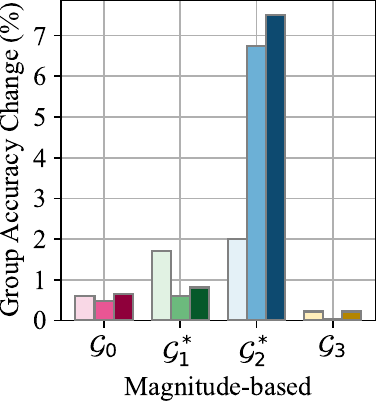}
    \includegraphics[width=0.25\textwidth]{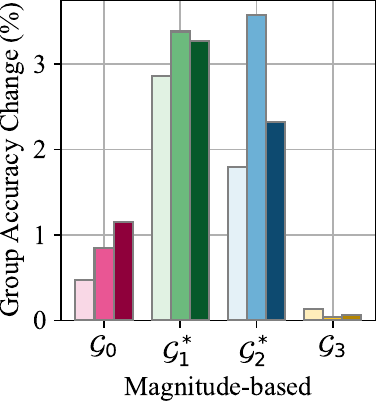}
    \includegraphics[width=0.25\textwidth]{figs/random_init_largest_magnitude_0.15_barplot.pdf}
    \caption{\textbf{Group accuracy change by random initialize Top-$k$ neuron(s) with the heuristics of gradient (top row) and magnitude (bottom row).} Within each group, three bars with gradated hues indicate the accuracy shift after random initializing the top-1, top-2, and top-3 neurons with the largest gradient or magnitude, respectively. The initialization $\epsilon_i\sim \mathcal{N}(\mathbf{0}, \mathbf{\sigma}^2)$, where the $\sigma=0.005,0.01,0.02$ from left column to right column, respectively.}
    \label{fig:appendix_random_init}
\end{figure}

\begin{figure}
    \centering
    \includegraphics[width=0.25\textwidth]{figs/random_noise_largest_activation_0.005_barplot.pdf}
    \includegraphics[width=0.25\textwidth]{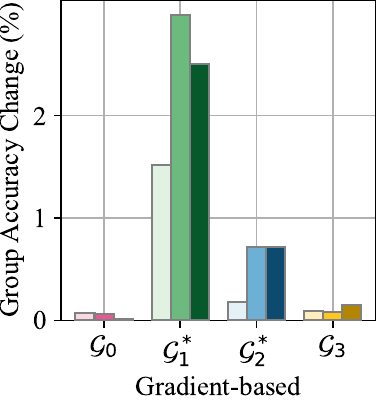}
    \includegraphics[width=0.25\textwidth]{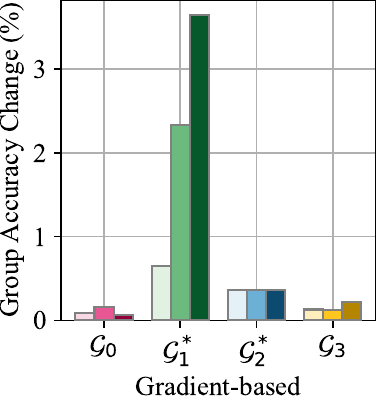}
    \includegraphics[width=0.25\textwidth]{figs/random_noise_largest_magnitude_0.005_barplot.pdf}
    \includegraphics[width=0.25\textwidth]{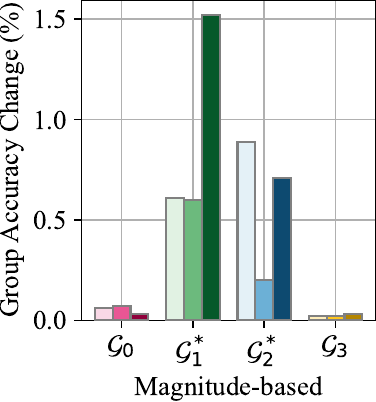}
    \includegraphics[width=0.25\textwidth]{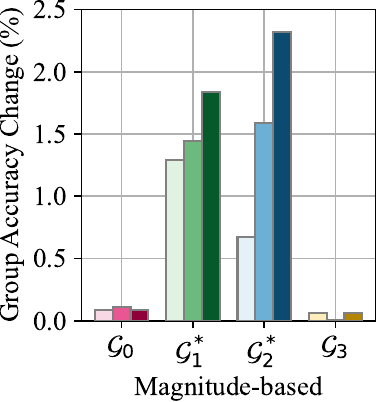}
    \caption{\textbf{Group accuracy change by adding random noise to Top-$k$ neuron(s)  with the heuristics of gradient (top row) and magnitude (bottom row).} Within each group, three bars with gradated hues indicate the accuracy shift after random initializing top-1, top-2, and top-3 neurons with the largest gradient or magnitude, respectively. The added noise $\epsilon_i\sim \mathcal{N}(\mathbf{0}, \mathbf{\sigma}^2)$, where the $\sigma=0.005,0.01,0.02$ from left column to right column, respectively.}
    \label{fig:appendix_random_noise}
\end{figure}

\end{document}